\documentclass[pmlr,twocolumn]{jmlr}

\usepackage{longtable}

\usepackage{booktabs}

\usepackage[load-configurations=version-1]{siunitx}

\theorembodyfont{\upshape}
\theoremheaderfont{\scshape}
\theorempostheader{:}
\theoremsep{\newline}

\newcommand{\change}{\textcolor{black}}
\newcommand{\new}{\textcolor{black}}
\usepackage{color,soul}

\newcommand{\beginappendix}{%
        \setcounter{table}{0}
        \renewcommand{\thetable}{S\arabic{table}}%
        \setcounter{figure}{0}
        \renewcommand{\thefigure}{S\arabic{figure}}%
     }

\title[Heterogeneous Treatment Effects in Multiple Outcomes]{Identifying Heterogeneous Treatment Effects in \titlebreak Multiple Outcomes using Joint Confidence Intervals}

  \author{\Name{Peniel N. Argaw\nametag{$^1$\thanks{Equal contribution}}} \Email{peniel@g.harvard.edu}\\
  \Name{Elizabeth Healey\nametag{$^{2,3}$$^*$}} \Email{ehealey@mit.edu}\\
  \Name{Isaac S. Kohane\nametag{$^3$}} \Email{isaac\_kohane@hms.harvard.edu}\\
  \addr $^1$Harvard John A. Paulson School of Engineering and Applied Sciences, Cambridge, MA \\
  \addr $^2$Massachusetts Institute of Technology, Cambridge, MA \\
  \addr $^3$Harvard Medical School, Boston, MA}

\begin{document}

\maketitle

\begin{abstract}
Heterogeneous treatment effects (HTEs) are commonly identified during randomized controlled trials (RCTs). Identifying subgroups of patients with similar treatment effects is of high interest in clinical research to advance precision medicine. Often, multiple clinical outcomes are measured during an RCT, each having a potentially heterogeneous effect. Recently there has been high interest in identifying subgroups from HTEs, however, there has been less focus on developing tools in settings where there are multiple outcomes. In this work, we propose a framework for partitioning the covariate space to identify subgroups across multiple outcomes based on the joint CIs. We test our algorithm on synthetic and semi-synthetic data where there are two outcomes, and demonstrate that our algorithm is able to capture the HTE in both outcomes simultaneously.
\end{abstract}

\begin{keywords}
causal inference, heterogeneous treatment effect, joint CIs, subgroup analysis \\
\end{keywords}

\section{Introduction}
\label{sec:intro}
The goal of clinical trials is to evaluate the efficacy of interventions, often by comparing the outcomes between treatment and control therapies. The efficacy is assessed through estimating the treatment effect. HTEs can explain the variability of treatment effects in a population over a covariate space by defining a set of subgroups with similar treatment effects, and these subgroups can then be analyzed in ways to advance precision medicine \citep{Varadhan2013}. For example, subgroup analysis can provide insight about which types of patients may respond exceptionally well or poorly to a given therapy \citep{Rekkas2020}.

In many cases, researchers and clinicians are interested in a given therapy's treatment effect on multiple outcomes. While there has been significant interest in developing techniques to discover HTEs from RCT data, they primarily focus on settings where there is just one outcome of interest. These do not capture the complexities of real-world scenarios where the therapy causes multiple effects \citep{Berkey1996}. Certain chronic diseases call for management through treatments that affect multiple clinical endpoints. Clinical trials often evaluate the treatment effect on primary and secondary outcomes, and it is common for clinical trials to have multiple primary outcomes \citep{Vickerstaff2015}. 

Clinicians, researchers, and pharmaceutical companies can benefit from knowing subgroup characteristics across multiple endpoints. For instance, a certain medication may show high efficacy in two subgroups, but show adverse side effects in one of the groups. Hence, it is critical that computational research focuses attention on the development of methods to analyze RCT data that captures multiple clinical endpoints.

One of the challenges in studying HTEs on multiple different outcomes is ensuring robust multivariate treatment effect estimates among subgroups. In the single outcome setting, recent work has explored subgroup identification in order to optimally separate treatment effects \citep{Rekkas2020}, and others identified subgroups by optimizing for intra- and inter-group variation and confidence intervals \citep{Lee2020-zd}. Algorithms that identify subgroups without accounting for intra-group variation run the risk of having subgroups with large CIs. This depreciates the robustness of the treatment effect estimate within a group and can limit the clinical utility of the findings. This can be further complicated when evaluating on multiple outcomes.

In our work, we offer a solution to this problem by proposing a novel method for partitioning treatment effects using the joint CIs of multiple outcome variables. We extend upon the partitioning framework from \cite{Lee2020-zd} by generating joint CIs using conformal prediction and quantile regression \citep{Lei2018,Romano2019_cqr,Lei2021}. We evaluate our approach on synthetic and semi-synthetic datasets inspired by clinical data and generalize our algorithm for multiple outcomes (specifically two outcomes in this study). We refer to our method as Multiple Outcome Partitioning using Joint Confidence Intervals, MOP-JCI. \\

\textbf{Our Key Contributions}
\begin{enumerate}
    \item We extend upon a framework for partitioning the covariate space to identify a set of subgroups with similar treatment effects across multiple outcomes using joint CIs.
    \item We deploy and evaluate a quantile individual treatment effect (ITE) estimator in the partitioning algorithm.
    \item We evaluate our approach on synthetic and semi-synthetic RCT datasets and show the robustness of our method on datasets containing correlation and heteroskedasticity.
\end{enumerate}

\section{Related Work}
\subsection{Subgroup Analyses}
Subgroup analysis is a common approach to identifying heterogeneities in treatment effect. Methodologies to estimate the heterogeneity include statistical tests \citep{Assmann2000,Alosh2015}, Bayesian modeling \citep{Jones2011,Pennello2018} and recursive partitioning \citep{Su2009,Athey2016,Lee2020-zd,Seibold2016-cu}. \cite{Lee2020-zd} proposed a confidence criterion for use in recursive partitioning, derived from the CIs of any mean ITE estimator, to ensure homogeneity within subgroups. These approaches, however, are all limited to a single outcome variable.

\subsection{Multiple Outcomes}
Multiple outcomes are common in RCTs and recent work has considered analyzing treatment effects in the setting of multiple outcomes. \cite{Kennedy2019} presented an approach to estimate the effects of multiple outcomes using a common scale, and discussed the dependency of treatment effects on covariates. \cite{Wu2022-oz} proposed a personalized policy generation method in the setting of multiple outcomes by weighting the treatment effect of each outcome. \cite{Yao2022-jy} proposed a method of treatment effect estimation that utilizes data across multiple outcomes. \cite{Yoon2011} evaluates likelihood-based methods to jointly test treatment effects across multiple outcomes. These approaches, however, do not focus on identifying subgroups where each group of patients show similar characteristics across all outcomes.


\section{Methods}
We begin by defining a preliminary framework which we use to estimate treatment effect. Next, we discuss the CI generation techniques used in our implementation. We then describe joint CI estimation for multiple outcomes and show the integration into a recursive partitioning algorithm to construct subgroups.

\subsection{Preliminaries}
We consider a setup of a RCT where there are two different outcomes of interest, outcome $A$ and outcome $B$. Namely, we have a total of $N$ samples each with covariates $X_i$, treatment assignment $t_i$, and outcome variables $Y^A_i$ and $Y^B_i$ for $i=1,...,N$. Here, $t_i$ is a binary variable $\{0,1\}$ representing the treatment group assignment for the sample. The outcome variables, $Y^A_i$ and $Y^B_i$, are scalar, continuous values for outcomes A and B, respectively. Our goal is to determine the ITE for each outcome as defined by $\mathbb{E}[Y^A_i(1)-Y^A_i(0)|X=x]$ and $\mathbb{E}[Y^B_i(1)-Y^B_i(0)|X=x]$ where $Y_i(1)$ and $Y_i(0)$ are the potential outcomes for each sample $i$ had they been treated with $1$ or $0$, respectively.

The ITE estimate for each outcome is the difference of the two regression models for the control and treated, $\hat{\mu}_0(x)$ and $\hat{\mu}_1(x)$ respectively. The regression models for outcomes A and B are defined as $\hat{\mu}^A_0(x)= \mathbb{E}[Y^A(0)|X=x]$, $\hat{\mu}^A_1(x)= \mathbb{E}[Y^A(1)|X=x]$ , $\hat{\mu}^B_0(x)= \mathbb{E}[Y^B(0)|X=x]$ , $\hat{\mu}^B_1(x)= \mathbb{E}[Y^B(1)|X=x]$.

CIs can be generated from each of the regressors using split conformal regression (SCR) \citep{Lei2018}. SCR initially splits the samples into two equal-sized sets, a training set $I_{tr}$ and a validation set $I_{val}$, then trains a regressor $\hat{\mu}^{I_{tr}}$, calculates the residuals of the trained model on the validation set $I_{val}$. The resulting CI bounds can be defined as $\hat{C}(x) = \left[ \hat{\mu}^{I_{tr}}(x)- \hat{Q}_{1-\alpha}^{I_{val}} \ , \ \hat{\mu}^{I_{tr}}(x) + \hat{Q}_{1-\alpha}^{I_{val}} \right]$ where $\hat{Q}_{1-\alpha}^{I_{val}}$ is the $(1-\alpha)(1+\frac{1}{|I_{val}|})$-th quantile of the residuals $\{|y_i - \hat{\mu}^{I_{tr}}(x_i)|\}_{i \in I_{val}}$, and $\alpha$ is the miscoverage rate used to ensure a coverage guarantee for each outcome $y$ such that $\mathbb{P}[y \in \hat{C}] \geq 1 - \alpha$.

\subsection{Split conformal quantile regression}

 Split conformal quantile regression (SCQR) is an alternative approach to estimate the ITE CIs for each regression model, $\hat{\mu}_0(X)$ and $\hat{\mu}_1(X)$, as described in \cite{Romano2019_cqr}. Again, we first split the data equally into a training and validation set, $I_{tr}$ and $ I_{val}$. The training set is used to fit the two quantile regression models, $\hat{q}_{\alpha}^{hi}$ and $\hat{q}_{\alpha}^{low}$ for a miscoverage rate $\alpha$. Using these estimators, we compute the calibration scores $E_{i}$ for each $i \in I_{val}$ by $E_{i} = \text{max}\{\hat{q}_{\alpha}^{low}(x_i)-Y_i, Y_i - \hat{q}_{\alpha}^{hi}(x_i)\}$.
The CIs of the estimator are $\hat{C}(x) = \left[ \hat{q}_{\alpha}^{low}(x)-\hat{Q}_{1-\alpha}^{I_{val}}(E) \ , \\ \hat{q}_{\alpha}^{hi}(x) + \hat{Q}_{1-\alpha}^{I_{val}}(E) \right]$, where $\hat{Q}_{1-\alpha}^{I_{val}}$ is the $(1-\alpha)(1+\frac{1}{|I_{val}|})$-th quantile of $\{|E_i|\}_{i \in I_{val}}$. 

\subsection{Joint CIs for ITE estimate}\label{sec:joint_ci_treat_control}
We calculate the conformalized ITE intervals by jointly considering the treated regression and control regression. We use the naive approach outlined in \cite{Lei2021} which directly compares the two intervals adjusts the miscoverage rate by dividing $\alpha$ by 2.  Accordingly, we define the CIs for the treated population as $[\hat{C}_{\alpha/2}^{low}(1;x)\ , \ \hat{C}_{\alpha/2}^{hi}(1;x)]$ and for control population as $[\hat{C}_{\alpha/2}^{low}(0;x),\hat{C}_{\alpha/2}^{hi}(0;x)]$ each with a coverage of $1 - \alpha/2$. The CI of the ITE estimator is defined as $\hat{C}_{ITE}(x) = [\hat{C}_{\alpha/2}^{low}(1;x)-\hat{C}_{\alpha/2}^{hi}(0;x)$, $\hat{C}_{\alpha/2}^{hi}(1;x)-\hat{C}_{\alpha/2}^{low}(0;x)]$.

\subsection{Joint CIs for multiple outcomes}
In the single outcome case, coverage is guaranteed for each outcome $y$ such that $\mathbb{P}[y \in \hat{C}] \geq 1 - \alpha$, where $\alpha$ was the miscoverage rate of the ITE estimate. We apply the Bonferroni correction to our coverage term in order to adjust the CIs for each outcome and guarantee a specified overall coverage across all the outcomes. This is done by taking the joint probability that each outcome's CIs are within a given coverage. Concretely, we divide the miscoverage rate by $d$, where $d$ is the total number of outcomes. Combining this adjustment with the previous adjustment in section \ref{sec:joint_ci_treat_control}, we set the  miscoverage rate as $\frac{\alpha}{2d}$ for each treated and control regressor, thus ensuring $1-\alpha$ coverage across the treatment effect of all outcomes ($\mathbb{P}[y \in \hat{C}_{ITE}] \geq 1 - \frac{\alpha}{2d}$).

\subsection{Recursive Partitioning Algorithm}
\new{We build upon the robust recursive partitioning algorithm (R2P) proposed \cite{Lee2020-zd} to partition the data into subgroups based on the covariate space}. We adapt their confidence criterion for use in the setting of multiple outcomes. They define a confidence criterion that aims to maximize heterogeneity across the subgroups and maximize homogeneity within the subgroups. They do this by minimizing the expected CI width $W_g$, with the expected absolute deviation $V_g$ within a group $g$. The expected width $W_g$ is defined as $\mathbb{E}[|\hat{C}_{g}(x)|]$, and the deviation $V_g = \mathbb{E}[\hat{v}_g(x)]$ where $v_g = (\hat{\mu}_{g}^{mean} -\hat{\mu}_{l}^{up}(x) \ \mathbb{I} \left[\hat{\mu}_{l}^{mean} >\hat{\mu}_{l}^{up}(x)\right] + (\hat{\mu}_{g}^{low}(x) - \hat{\mu}_{g}^{mean}) \ \mathbb{I} \left[\hat{\mu}_{l}^{mean} <\hat{\mu}_{g}^{low}(x)\right]$.

The expected absolute deviation $V_g$ within a group $g$ can otherwise be explained as the error between the the CI bound and the average outcome. Together, the partitioning is done with the following objective. 
\begin{equation*}
   \textrm{minimize} \sum_{g \in \Pi} \lambda W_g + (1 - \lambda) V_g
\end{equation*}
where $\lambda$ is a hyperparameter used to vary the weight on $V_g$ and $W_g$ and $\Pi$ is the set of partitions. We extend this criterion to work with multiple outcomes by summing the regions for each outcome, using predefined weights. In the two outcome case, the objective function is as follows:
\begin{equation*}
\begin{split}
   \textrm{minimize} \sum_{g \in \Pi} \lambda (\beta W^A_g+(1-\beta)W^B_g) + \\ (1 - \lambda) (\beta V^A_g+(1\change{-}\beta\change{)} V^B_g)
\end{split}
\end{equation*}
Here, $\beta$ is a tuning parameter to weight the outcomes. It can be tuned to give preference for one outcome over another, and to account for differences in expected magnitude of the outcomes. We provide a modified objective function for more than two outcomes in Appendix \ref{app:methods}.

For the SCQR method, we simplify the objective function to the setting where $\lambda$ is 0. This is because in the SCQR setting, the CIs for each covariate are determined by the quantile estimator. They do not change with further calibration after each split. Thus, the objective function for the SCQR method in the setting of two outcomes is defined as:
\begin{equation*}
   \textrm{minimize} \sum_{g \in \Pi}  (\beta V^A_g+(1-\beta \change{)} V^B_g)
\end{equation*}

We further adapt the robust recursive partitioning algorithm proposed in \cite{Lee2020-zd} to partition on two outcomes and to work with a quantile estimator in Algorithm \ref{algo_partition} (the partitioning algorithm for SCR can be found in the Appendix \ref{app:methods}). 

\begin{algorithm2e}
\caption{SCQR Recursive Partitioning on Two Outcomes}
  \KwIn{$G_{node}$, data} 
    \For{Covariate $j$ in data}{
        \For{Unique value $x$ of covariate $j$}{
            Split data in two branches on $x$\\
            Compute $V^A$ for each branch \\
            Compute $V^B$ for each branch \\
            Set $ G_{split} $ to $G_{node}$ less the sum of $ (\beta V^A+(1-\beta)V^B)$ from each branch\\
        }
    }
    Save best $G_{split}$ with covariate $j$ \\
    \uIf{$ G_{split}> \gamma G_{node}$ }{
        $ G_{node} \leftarrow G_{split}$ \\
        Partition on each branch \\
    }
    \Else{
        Set current node to leaf\\
    }
\label{algo_partition}
\end{algorithm2e}

To work with the quantile estimator, we first perform SCQR on the training data using the $I_{tr}$ to fit the random forest quantile estimator and $I_{val}$ to compute the confidence metrics. We then compute $\hat{C}^A_{ITE}$ and $\hat{C}^B_{ITE}$ from $I_{val}$ of each outcome. Since the SCQR estimates the quantile distribution of the treatment effect across the covariate space, there is no need for recalibrating during the partitioning, reducing computational burden.

We use the $I_{val}$ data to partition the data using a recursive function. We start with the entire set  $I_{val}$ as the root node and use Algorithm \ref{algo_partition} to create nodes by computing the a joint confidence score using $V_g$. The recursive function takes in a node and the $\hat{C}_{ITE}$ of each estimator to calculate the $V_g$. The best gain is computed for each split along a single covariate as $G_{split}$. A node is split into branches when the candidate split, $G_{split}$, is greater than $\gamma G_{node}$, where $\gamma$ controls for regularization of the number of subgroups. The resulting leaves make up the subgroups.

\section{Experimental Design}

\subsection{Experiments}

We generate a set of experiments to evaluate our method and the use of the two different conformal regression techniques.
\begin{enumerate}
    \item Baseline (R2P): We use R2P to partition on a single outcome at a time. We observe the subgroup formation from partitioning on each outcome separately. 
    \item Our method (MOP-JCI)
    \begin{itemize}
        \item \new{SCR} approach: We use SCR to generate joint CIs using the Bonferroni correction to guarantee joint coverage of both outcomes. We partition using the CI regions of both outcomes in the minimization function.
        \item \new{SCQR} approach:  We use SCQR to generate joint CIs using the Bonferroni correction to guarantee joint coverage of both outcomes. We partition using the CI regions of both outcomes in the minimization function.
    \end{itemize}

\end{enumerate}

Moreover, we experimented on various ITE estimators. These estimators included Causal Multi-task Gaussian Process (CMGP) \citep{Alaa2017}, Random Forest (RF),  and Quantile Random Forest (QRF). For CMGP, we use the implementation provided by the authors. RF and QRF are all adapted from \textit{scikit-learn} and \textit{scikit-garden} to estimate confidence bounds across the treated and control populations for multiple outcomes.  \new{See Appendix \ref{app:ite} and \ref{app:methods} for details on hyperparameter tuning}. Additionally, this study is based on the assumptions listed in Appendix \ref{app:methods}. Our implementation is available at \href{https://github.com/pargaw/MOP-JCI}{\texttt{https://github.com/pargaw/MOP-JCI}}. 

\subsection{Evaluation Metrics}
We evaluate the statistical significance of the defined subgroups and the precision of the regressors. As our goal was to maximize heterogeneity across groups and homogeneity within groups, we evaluated the variance found within and across the groups. Variance across the groups was defined as $V_{across} = Var ( {Mean(S_{g}^{test})}^{G}_{g=1})$ where $S_{g}^{test}$ is the set of test samples in group $g$ and $G$ is the total number of subgroups. Variance within the groups was defined as $V_{within} (S^{test}) = \frac{1}{G} \sum_{g=1}^{G} Var(S_{g}^{test})$. Further, we evaluate the true coverage of the CIs generated (Cov) by computing the percentage of time that both outcome predictions fall within the CI with miscoverage set at $0.1$. We report the mean width of the CIs (CI Width) as well.  Additionally, we evaluated the error of our ITE estimators using the precision in estimation of heterogeneous effect (PEHE) \citep{Hill2011}.

In our experiments, we run each algorithm 30 times to take the mean and standard deviation of the metrics. We additionally compute the percentage of iterations that the partitioning algorithm split the subgroups on the expected covariates (Split Acc). Similarly, we compute the percentage of iterations where the algorithm split on unexpected covariates,  (Split Err). Unexpected covariates are covariates that have little or no underlying effect on the outcome distribution. These metrics are added to ensure the subgroups are formed based on ground truth knowledge from the data generation. 

\subsection{Datasets}\label{datasets}
We evaluate our work on synthetic and semi-synthetic datasets, where each have two outcomes. Additionally, to assess the robustness of our approach, we evaluate our model on variations of our synthetic dataset that exhibit uncorrelated covariates, correlated covariates, and heteroscedasticity. More details on the outcome generation and distributions for each of the datasets can be found in Appendix \ref{app:datasets}.

\subsubsection{Synthetic Data}
We adapt synthetic data proposed in \cite{Lee2020-zd} to represent multiple outcomes. The synthetic data was inspired by the clinical trial of remdesivir on COVID-19 \citep{Beigel2020}. The synthetic data consists of simulated versions of covariates used in the trial, with values randomized on varying distributions (see Appendix \ref{app:datasets}). The outcome in the synthetic data is days to clinical improvement, with data simulated to show the relationship between faster clinical improvement and shorter time from symptom onset to starting the trial. We extend the synthetic data to include a second outcome reflective of an adverse event in the trial: alanine aminotransferase (ALT) levels on the last day of the trial (day 28). We focus on the version of this data where the outcomes are uncorrelated. Research has suggested that high ALT levels at baseline may put patients at increased risk for liver function deterioration from remdesivir \citep{Charan2021}. We simulate increased risk of higher end-point ALT as a function of baseline ALT. We simulate data from a trial where the time to improvement is measured as efficacy, and liver function deterioration is measured as an adverse event.  In the primary version of this synthetic data, the outcomes are uncorrelated.  Additional variations of this dataset are described and evaluated in Appendix \ref{app:datasets} and \ref{app:results}. 

\subsubsection{Semi-synthetic Data}
For the semi-synthetic case, we used the Infant Health Development Program (IHDP) dataset. The dataset was inspired by a real RCT where the goal was to evaluate the efficacy of early intervention to improve the health and development of low-birth-weight, premature infants \citep{Gross1993}. The deidentified covariates were extracted from the original study, and the outcomes were simulated using the Response Surface B described in \cite{Hill2011}. The dataset was adapted for a two-outcome setting by choosing a different covariate to relate to each outcome (Neo-natal health index (nnhealth) and mom age respectively). These outcomes and covariate relationships are inspired by the results found in \cite{Baumeister1996}.

\section{Results}\label{results}

\subsection{Synthetic Data}
\begin{figure*}[ht!] 
  \centering
  (a){\includegraphics[scale=.3,trim={0cm 0cm 0cm 0cm}]{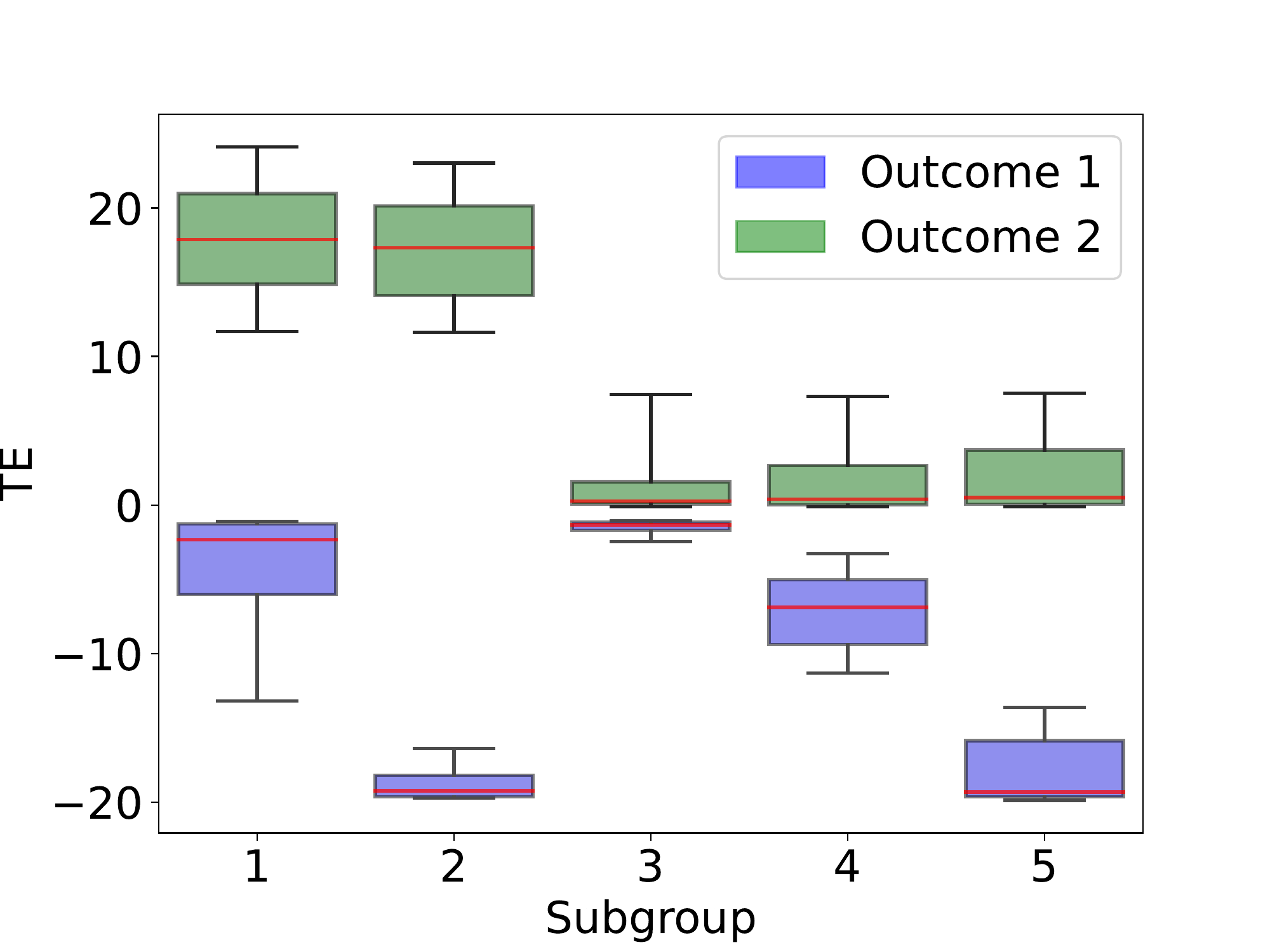}}
  (b){\includegraphics[scale=.3,trim={0cm 0cm 0cm 0cm}]{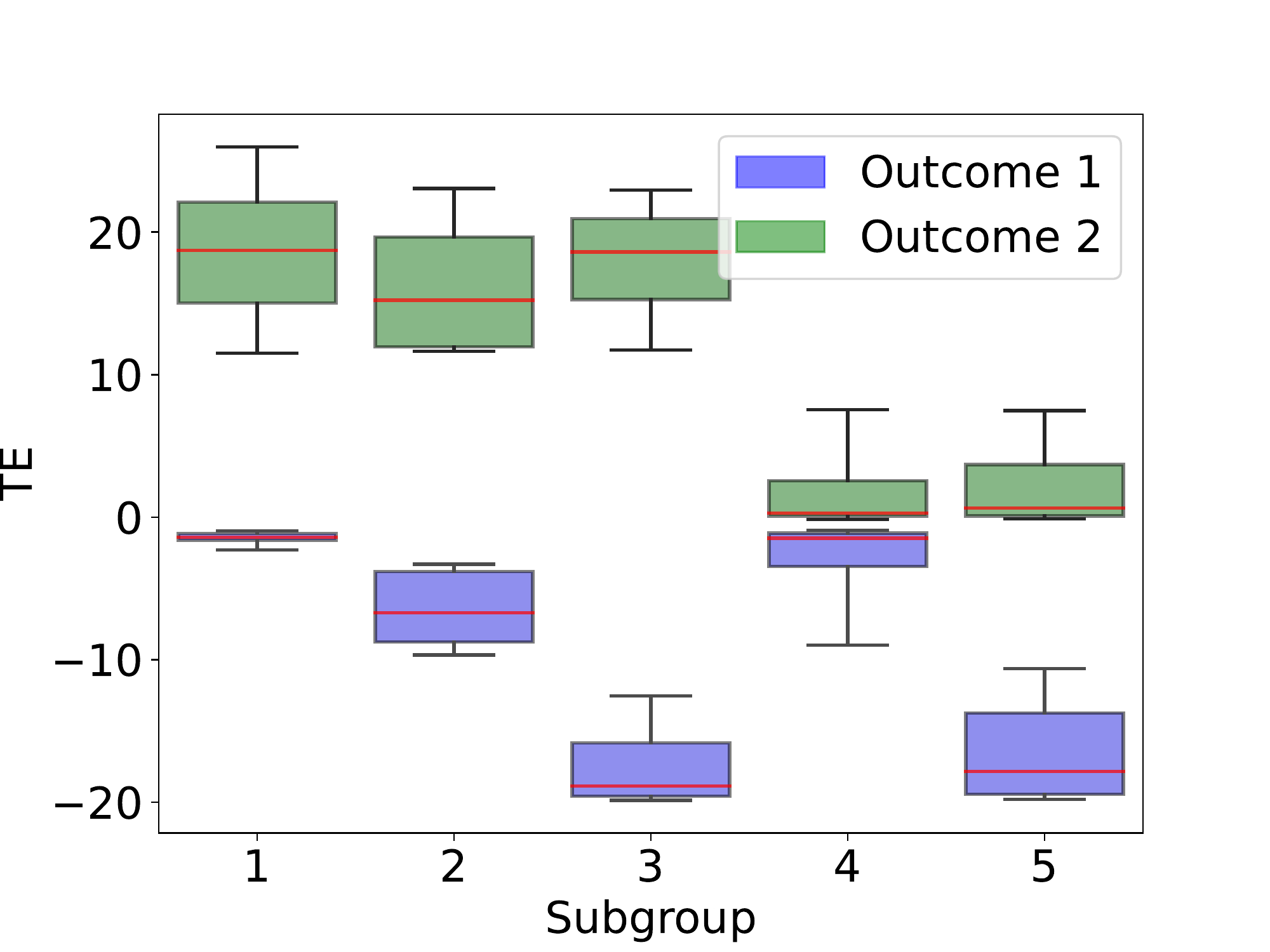}}
  \caption{\textbf{Jointly partitioned subgroups on synthetic data (MOP-JCI).} Subgroups defined in each outcome when partitioned using the RF estimator on the SCR method (a) and the SCQR method (b). Whiskers show the 25th and 75th percentiles of the treatment effect (TE).}
  \label{fig:uncorr_joint}
  \vspace{-1.5em}
\end{figure*}

\begin{table*}[t!]
\scriptsize
\centering
\setlength{\tabcolsep}{1pt}
\begin{tabular}{p{3cm}|p{1cm}|p{1cm}p{1cm}p{1cm}p{1cm}p{1cm}|p{1cm}p{1cm}p{1cm}p{1cm}p{1cm}}
\toprule
\multicolumn{12}{l}{Baselines (R2P)} \\\hline
&  & \multicolumn{5}{c|}{Outcome 1} & \multicolumn{5}{c}{Outcome 2} \\\hline
&   Num groups &  $V_{across}$ &   $V_{within}$ &   PEHE &  CI Width &   Cov        &  $V_{across}$ &   $V_{within}$ &   PEHE &  CI Width  & Cov \\\hline
CMGP on outcome 1 &    5.50 ±0.19 &  51.07 ±1.63 &    \textbf{1.64 ±0.11} &    0.20 ±0.02  &    1.22 ±0.13 &   98.62 ±0.47 &   3.09 ±1.60 &   74.53 ±1.70 &               - &               - &               - \\
CMGP on  outcome 2 &    5.03 ±0.18 &   4.13 ±3.10 &   52.73 ±2.98 &               - &               - &               - &  68.97 ±2.07 &    \textbf{4.28 ±1.19} &    0.62 ±0.16  &    3.83 ±1.12 &   97.97 ±0.59 \\
RF on  outcome 1 &    5.50 ±0.21 &  \textbf{52.51 ±1.65} &    1.90 ±0.20 &    0.58 ±0.04  &    4.13 ±0.37 &   98.58 ±0.64 &   1.96 ±0.48 &   76.63 ±1.73 &               - &               - &               - \\
RF on outcome 2 &    4.90 ±0.18 &   3.77 ±2.23 &   52.79 ±2.02 &               - &               - &               - &  \textbf{69.49 ±2.44} &    7.07 ±1.04 &    1.08 ±0.10  &    9.97 ±1.01  &   98.62 ±0.73 \\
\bottomrule
\end{tabular}

\setlength{\tabcolsep}{1pt}
\begin{tabular}{p{2cm}|p{1cm}p{1cm}p{1cm}|p{1cm}p{1cm}p{1cm}p{1cm}|p{1cm}p{1cm}p{1cm}p{1cm}|p{1cm}}
\toprule
\multicolumn{12}{l|}{Jointly Partitioned (MOP-JCI)} \\\hline
&  & & & \multicolumn{4}{c|}{Outcome 1} & \multicolumn{4}{c|}{Outcome 2} & 
\\\hline
&     Num groups & Split Acc &          Split Err &       $V_{across}$ &        $V_{within}$ &            PEHE &               CI Width &       $V_{across}$ &        $V_{within}$ &            PEHE &               CI Width &     Cov (joint) \\ \hline
CMGP (SCR) &    4.97 ±0.37 &        97\% &        13\% &   46.51 ±3.68 &     9.67 ±3.67 &     0.18 ±0.02 &     1.17 ±0.14 &   62.71 ±4.81 &    13.60 ±4.81 &     1.06 ±0.29 &     7.34 ±2.26 &    97.97 ±0.45 \\
RF (SCR) &   4.80 ±0.25 &      100\% &       17\% &  48.37 ±1.48 &    7.58 ±1.15 &    0.57 ±0.05 &    4.48 ±0.35 &  \textbf{65.98 ±2.13} &   \textbf{11.30 ±0.70} &    1.10 ±0.10 &   11.62 ±1.08 &   98.73 ±0.65 \\
QRF (SCQR) &   4.77 ±0.21 &      100\% &       10\% &  \textbf{48.92 ±0.94} &    \textbf{6.11 ±0.37} &    0.61 ±0.05 &    6.08 ±0.49 &  62.66 ±1.62 &   11.91 ±0.65 &    1.18 ±0.10 &   16.20 ±1.44 &   98.62 ±0.86 \\
\bottomrule
\end{tabular}
\caption{Results from synthetic data. We take the mean and standard deviation of each metric across 30 runs. Num groups is the number of subgroups generated. Best performance for $V_{across}$ and $V_{within}$ in each column are in bold.}
\vspace{-2.5em}
\label{tab:synth}
\end{table*}

In this section we show the results of the partitioning methods performed on the synthetic dataset. Table \ref{tab:synth} shows the baseline partitioning from R2P on a single outcome, and the MOP-JCI methods using the SCR and SCQR approach. Looking at the baseline R2P, when the partitioning is done on outcome 1, clear subgroups are identified with respect to outcome 1 noting that variance in each subgroup is low and variance across each subgroup is high. However, the characteristics of outcome 2 in the corresponding subgroups formed when only outcome 1 is partitioned are not well defined. Whereas, the MOP-JCI methods are able to capture low within group variance and high across group variance for both outcomes simultaneously in the same partition. All MOP-JCI methods are able to identify the expected covariates to split on (denoted by Split Acc). Figure \ref{fig:uncorr_joint} shows the subgroups formed by the MOP-JCI methods. We can see that in both of these cases, the subgroups are well-defined across both outcomes. See Figure \ref{fig:uncorr_scr_cmgp_sep} for the subgroup formation from R2P on a single outcome and Appendix \ref{app:subgroup_chars} for the full subgroup characteristics. The performance metrics at different values of the tuning parameter $\beta$ are found in Figure \ref{fig:beta synthetic1}. 

\subsection{Semi-synthetic Data}
Similarly in the IHDP data, we show the effects of partitioning separately on each outcome using R2P and jointly partitioning using SCR and SCQR using MOP-JCI.  Table \ref{tab:semisynth} shows the results from the partitioning algorithms. See Appendix Figure \ref{fig:ihdp_1_scr_cmgp_sep} and Figure \ref{fig:ihdp_1_joint} for the treatment effect of subgroup formations. The MOP-JCI methods are able to achieve similar variance across groups and similar variance within groups for each outcome as when they are partitioned separately. The error metrics when reporting the results of the IHDP data are rather high. This is due to the fact that the generation of the outcomes involves many covariates with a small effect. The accuracy metric is more useful here, since there is one dominant covariate contributing to each outcome effect. The performance metrics at different values of the tuning parameter $\beta$ are found in Figure \ref{fig:beta semi synthetic}. 

\begin{table*}[htbp!]
\scriptsize
\centering
\setlength{\tabcolsep}{1pt}
\begin{tabular}{p{3cm}|p{1cm}|p{1cm}p{1cm}p{1cm}p{1cm}p{1cm}|p{1cm}p{1cm}p{1cm}p{1cm}p{1cm}}
\toprule
\multicolumn{12}{l}{Baselines (R2P)} \\\hline
&  & \multicolumn{5}{c|}{Outcome 1} & \multicolumn{5}{c}{Outcome 2} 
\\\hline
&   Num groups &  $V_{across}$ &   $V_{within}$ &   PEHE &  CI Width &   Cov        &  $V_{across}$ &   $V_{within}$ &   PEHE &  CI Width  & Cov \\\hline
CMGP on outcome 1  &    4.17 ±0.17 &  \textbf{18.02 ±2.29} &   \textbf{21.51 ±1.51} &    2.59 ±0.19  &   14.12 ±1.34 &   95.38 ±1.21 &   0.09 ±0.04 &    0.81 ±0.04 &               - &               - &               - \\
CMGP on outcome 2 &    4.10 ±0.20 &   3.84 ±1.65 &   39.62 ±3.24 &               - &               - &               - &   \textbf{0.57 ±0.03} &    \textbf{0.27 ±0.02} &    0.27 ±0.03  &    1.70 ±0.20 &   97.93 ±0.51 \\
RF on outcome 1 &    2.47 ±0.29 &  11.85 ±2.63 &   29.22 ±2.78 &    4.06 ±0.20  &   26.07 ±2.47 &   97.64 ±0.68 &   0.01 ±0.01 &    0.84 ±0.04 &               - &               - &               - \\
RF on outcome 2 &    1.20 ±0.15 &   0.61 ±1.08 &   37.63 ±2.59 &               - &               - &               - &   0.03 ±0.04 &    0.82 ±0.05 &    0.66 ±0.03  &    4.20 ±0.20 &   99.38 ±0.31 \\
\bottomrule
\end{tabular}

\setlength{\tabcolsep}{1pt}
\begin{tabular}{p{2cm}|p{1cm}p{1cm}p{1cm}|p{1cm}p{1cm}p{1cm}p{1cm}|p{1cm}p{1cm}p{1cm}p{1cm}|p{1cm}}
\toprule
\multicolumn{12}{l|}{Jointly Partitioned (MOP-JCI)} \\\hline
&  & & & \multicolumn{4}{c|}{Outcome 1} & \multicolumn{4}{c|}{Outcome 2} & \\\hline
&     Num groups & Split Acc &          Split Err &       $V_{across}$ &        $V_{within}$ &            PEHE &               CI Width &       $V_{across}$ &        $V_{within}$ &            PEHE &               CI Width &     Cov (joint) \\ \hline
CMGP (SCR) &   4.17 ±0.14 &       43\% &       87\% &  15.39 ±1.93 &   25.10 ±2.51 &    2.63 ±0.18 &   12.69 ±1.20 &   \textbf{0.33 ±0.08} &    \textbf{0.52 ±0.07} &    0.25 ±0.02 &    1.68 ±0.18 &   93.62 ±1.03 \\
RF (SCR) &   3.83 ±0.26 &       33\% &       90\% &  13.07 ±1.65 &   25.91 ±2.65 &    4.01 ±0.20 &   21.67 ±2.00 &   0.13 ±0.05 &    0.73 ±0.06 &    0.67 ±0.02 &    3.38 ±0.16 &   94.93 ±1.00 \\
QRF (SCQR) &   4.37 ±0.23 &       53\% &       87\% &  \textbf{17.60 ±2.04} &   \textbf{23.56 ±1.92} &    3.95 ±0.20 &   21.25 ±0.97 &   0.32 ±0.05 &    0.55 ±0.06 &    0.55 ±0.02 &    3.07 ±0.10 &   95.64 ±0.93 \\
\bottomrule
\end{tabular}
\caption{Results from semi-synthetic data. We take the mean and standard deviation of each metric across 30 runs. Num groups is the number of subgroups generated. Best performance for $V_{across}$ and $V_{within}$ in each column are in bold.}
\label{tab:semisynth}
\vspace{-2.5em}
\end{table*}

\section{Conclusion}\label{conclusion}
With RCT data being widely available, methods to properly analyze the data are essential in order to advance precision medicine. Our work introduces a method that can identify subgroups of patients in an RCT whose  response across multiple outcomes is homogeneous. By using a joint CI, we ensure that the subgroups have robust coverage across the multiple outcomes, regardless if the outcomes show similar or opposing treatment effects.

To evaluate our approach, we tested a baseline method that partitioned the covariate space solely on single outcomes, and demonstrated the shortfall by observing poor heterogeneity across subgroups for both outcomes. Our method showed that we can partition considering the variance across subgroups and within subgroups for each outcome simultaneously, as compared to when they are partitioned on each outcome separately. Additionally, we ensured the validity of our results by reporting the joint coverage and the percentage of correct and incorrect covariate splits. We showed how the tuning parameter used in our method can be used to favor one outcome over another. Our method paves the way for future work focusing on statistically advanced methods of analyzing RCT data where there are multivariate and multi-output effects. Lastly, we implemented a quantile ITE estimator to partition the data, which reduced the need for a tuning parameter in the algorithm and recalibration at each split, reducing computational burden.

\subsubsection*{Limitations}
\vspace{-0.5em}
Though we used semi-synthetic data with real-world covariates, we only evaluated our data on scenarios where the outcomes are synthetically generated. Future effort should be made towards evaluating the performance of our method on real-world outcomes. In the current approach, we use the Bonferroni correction to adjust the miscoverage rate for both the joint CIs on the ITE estimate for a single outcome, and the joint ITE estimate for multiple outcomes. This method for estimating joint CIs can be conservative, especially as the correlation between outcomes grows. Future work should focus on more precise joint CI estimation. Lastly, future work should investigate how performance changes when there are more than two outcomes. 

\vspace{-1em} 
\section{Broader Impacts and Ethics}\label{impact}

Our method has the potential to advance precision medicine by identifying subgroups where the response is homogeneous across multiple outcomes. Despite the potential for positive impact as a result of our work, we note a few potential ethics considerations. Primarily, the results of our work are not meant to be interpreted as definitive conclusions drawn about subsets of patients, but rather meant to allow clinicians to propose hypotheses for further investigation.  By identifying which covariates determine the subgroups, our method has the potential to serve as a tool to help researchers form testable hypotheses about which patients may be ideal candidates for a therapy. 

\acks{E.H. is supported by the National Science Foundation Graduate Research Fellowship under Grant No. 2141064.}

\bibliography{argaw22}

\newpage
\onecolumn
\twocolumn

\appendix
\beginappendix


\section{Datasets}\label{app:datasets}

\subsection{Synthetic Data}
This section describes the synthetic data that we adapted from \cite{Lee2020-zd}. We use a logistic function to generate both outcome distributions. The outputs were generated using the following equations to simulate for the control and treated populations, in the first outcome:
\begin{equation*}
\begin{split}
    \textrm{Control: } y_A(0) \sim  \\ \mathcal{N}(X_{0} \beta + 
    (1 + e^{-(x_j - m_j)})^{-1} + 20, 0.1) \\
    \textrm{Treatment: }  y_A(1) \sim \\ \mathcal{N}(X_{0} \beta + 20(1 + e^{-(x_j - m_j)})^{-1}, 0.1) \\
\end{split}
\end{equation*}
and the second:
\begin{equation*}
\begin{split}
    \textrm{Control: } y_B(0) \sim \\ \mathcal{N}(X_{0} \beta + (1 + e^{-(x_j - m_j)})^{-1} + x_j, 0.1) \\
    \textrm{Treatment: }  y_B(1) \sim \\ \mathcal{N}(X_{0} \beta + x_j(1 + e^{-(x_j - m_j)})^{-1} + x_j, 0.1) \\
\end{split}
\end{equation*}
where $m_j$ applies a shift by the mean of the covariate $x_j$, $X_0$ represents a matrix of all the covariate values except for the covariate $x_j$ and $\beta$ applies coefficients that are randomly sampled among $(0, 0.1, 0.2, 0.3, 0.4)$ probabilities $(0.6, 0.1, 0.1, 0.1, 0.1)$ respectively. Table \ref{tab:dist_synth_b} shows the distributions of the simulated covariates.

\begin{table*}[htbp!]
\centering
\begin{tabular}{ |l|l| } 
 \hline
 age & $\sim \mathcal{N}(66, 4.1)$  \\[0.1cm]
 white blood cell count (x$10^9$ per L) & $\sim \mathcal{N}(6.2, 1)$ \\[0.1cm]
 lymphocyte count (x$10^9$ per L) & $\sim \mathcal{N}(0.8, 0.1)$ \\[0.1cm]
 platelet count (x$10^9$ per L) & $\sim \mathcal{N}(183, 20.4)$ \\[0.1cm]
 serum creatinine (U/L) & $\sim \mathcal{N}(68, 6.6)$ \\[0.1cm]
 asparatate aminotransferase (U/L) & $\sim \mathcal{N}(31, 5.1)$ \\[0.1cm]
 alanine aminotransferase (U/L) & $\sim \mathcal{N}(16, 5.1)$ \\[0.1cm]
 lactate dehydrogenase (U/L) & $\sim \mathcal{N}(339, 51)$ \\[0.1cm]
 creatine kinase (U/L) & $\sim \mathcal{N}(76, 21)$ \\[0.1cm]
 time from symptom onset to starting the trial (days) & $\sim Unif(9, 14)$ \\[0.1cm]
 \hline
\end{tabular}
\caption{\label{tab:dist_synth_b}Distributions of covariates in synthetic data.}
\end{table*}

This dataset was modified to assess for the robustness of our model with two additional variations of this dataset. One showed the effect of correlated covariates, specifically time and ALT levels. This accordingly results in correlated outcomes.  The other showed a heteroscedastic trend for time on one outcome and ALT levels on the other. In all synthetic datasets, we created a set of $300$ and $200$ samples for training and testing respectively.

\begin{figure*}[htpb!]
\vspace{-4em} 
\centering
\includegraphics[scale=.36,trim={0cm 0cm 0cm 0cm}]{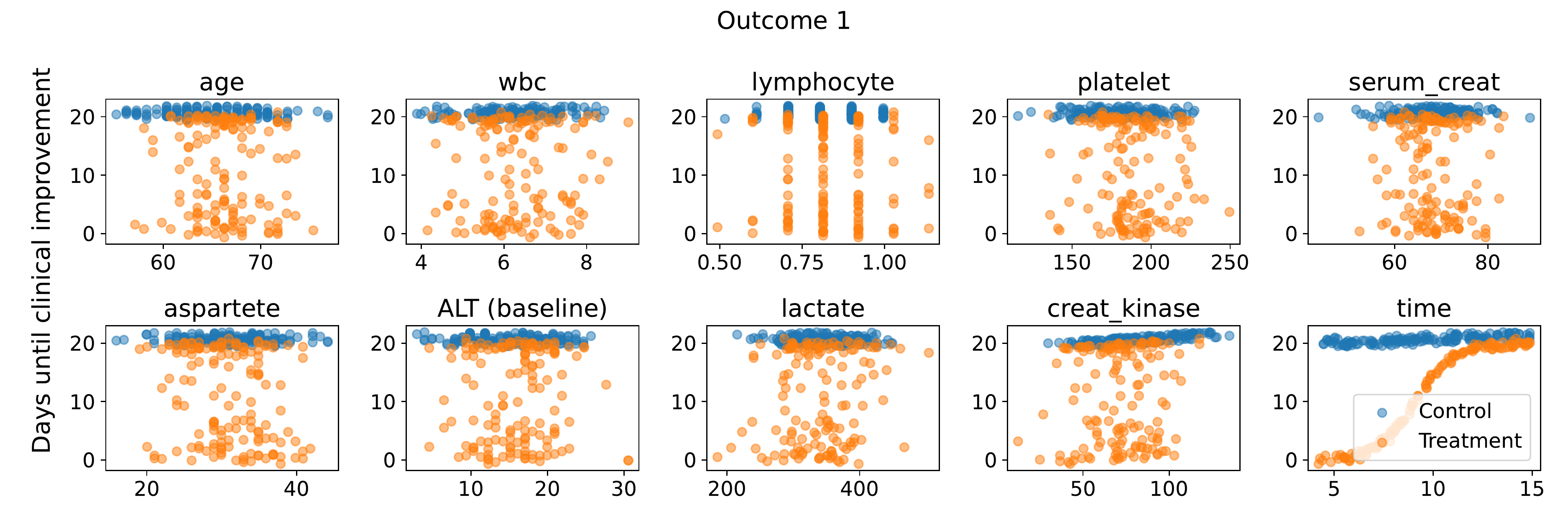}
\includegraphics[scale=.36,trim={0cm 0cm 0cm 0cm}]{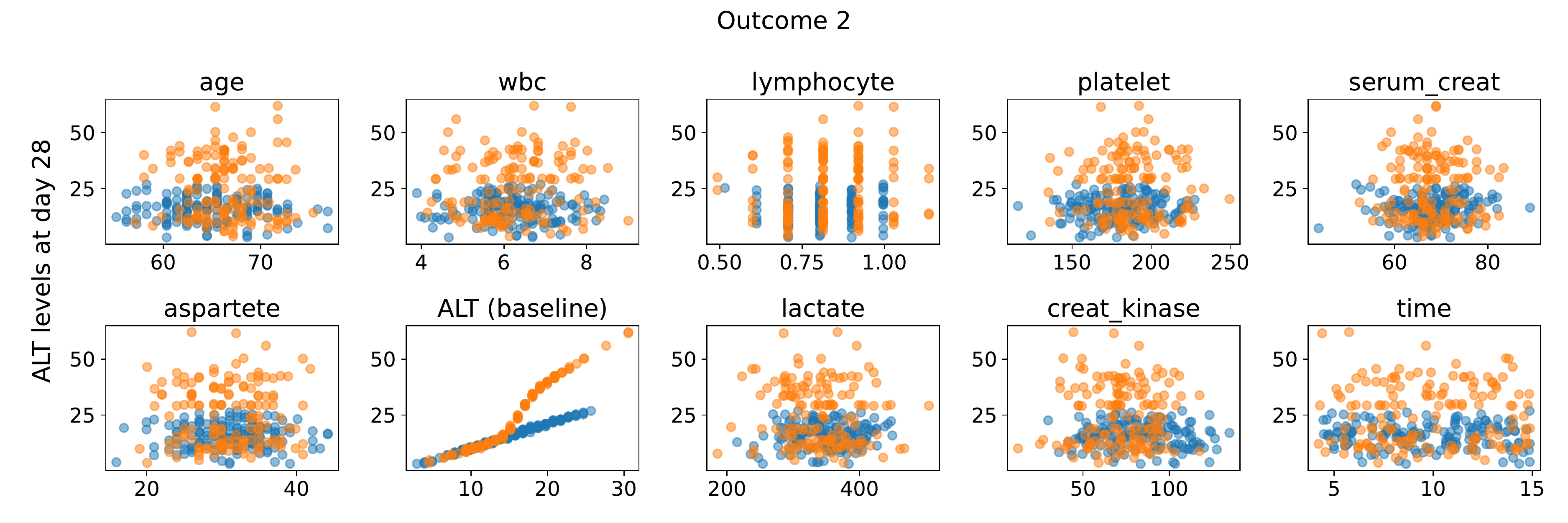}
\caption{\textbf{Treatment effect across covariates in the synthetic dataset.} Distributions are split across control and treatment populations, shown in blue and orange respectively.}
\label{fig:dist_synthb_uncorrout}
\end{figure*}

\begin{figure*}[htpb!]
\vspace{-4em} 
\centering
\includegraphics[scale=.36,trim={0cm 0cm 0cm 0cm}]{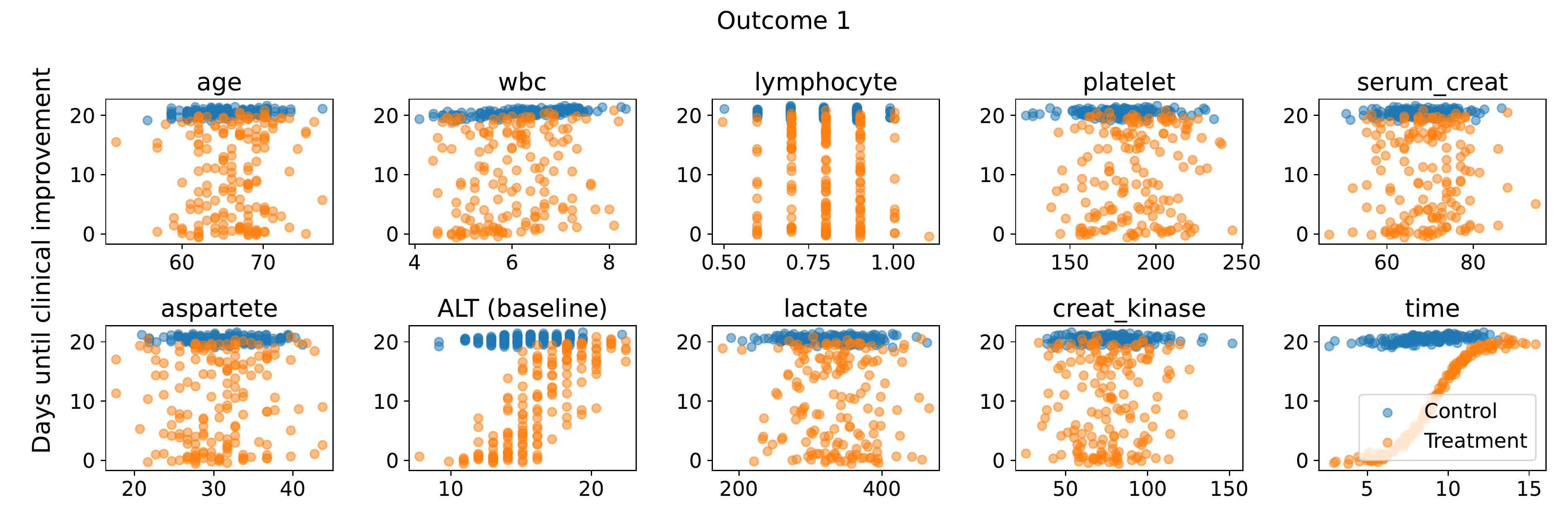}
\includegraphics[scale=.36,trim={0cm 0cm 0cm 0cm}]{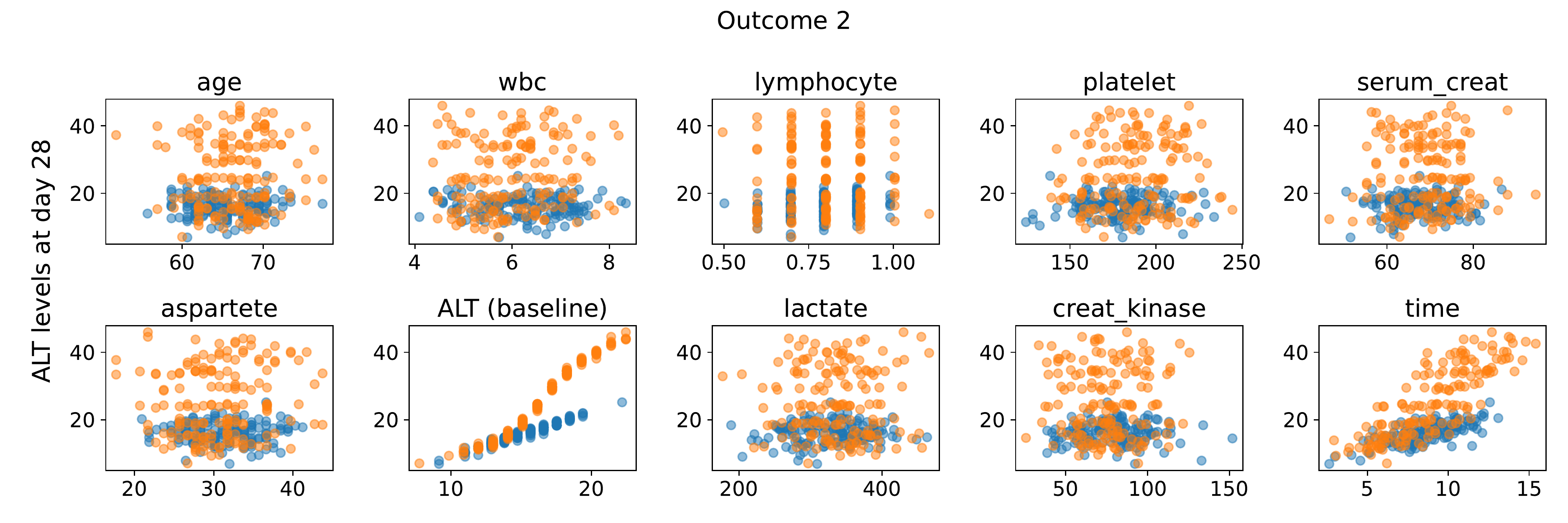}
\caption{\textbf{Treatment effect across covariates in the synthetic dataset, when there are correlated covariates.} Distributions are split across control and treatment populations, shown in blue and orange respectively.}
\label{fig:dist_synthb_corrcov}
\end{figure*}

\begin{figure*}[htpb!]
\centering
\includegraphics[scale=.36,trim={0cm 0cm 0cm 0cm}]{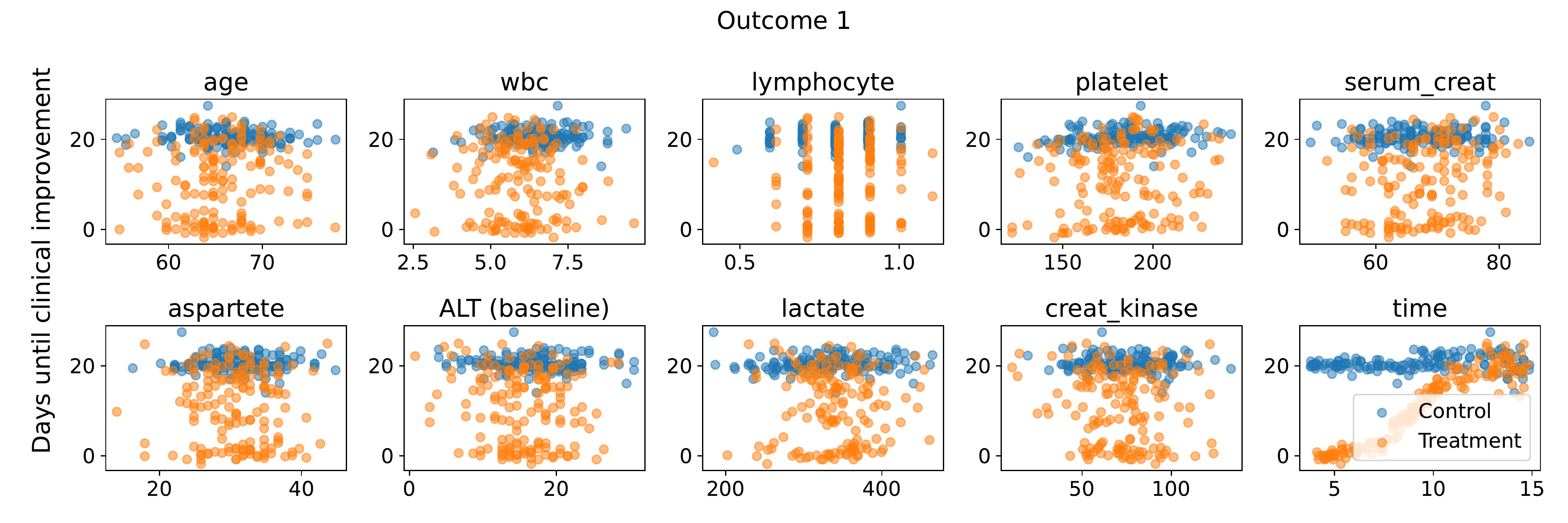}
\includegraphics[scale=.36,trim={0cm 0cm 0cm 0cm}]{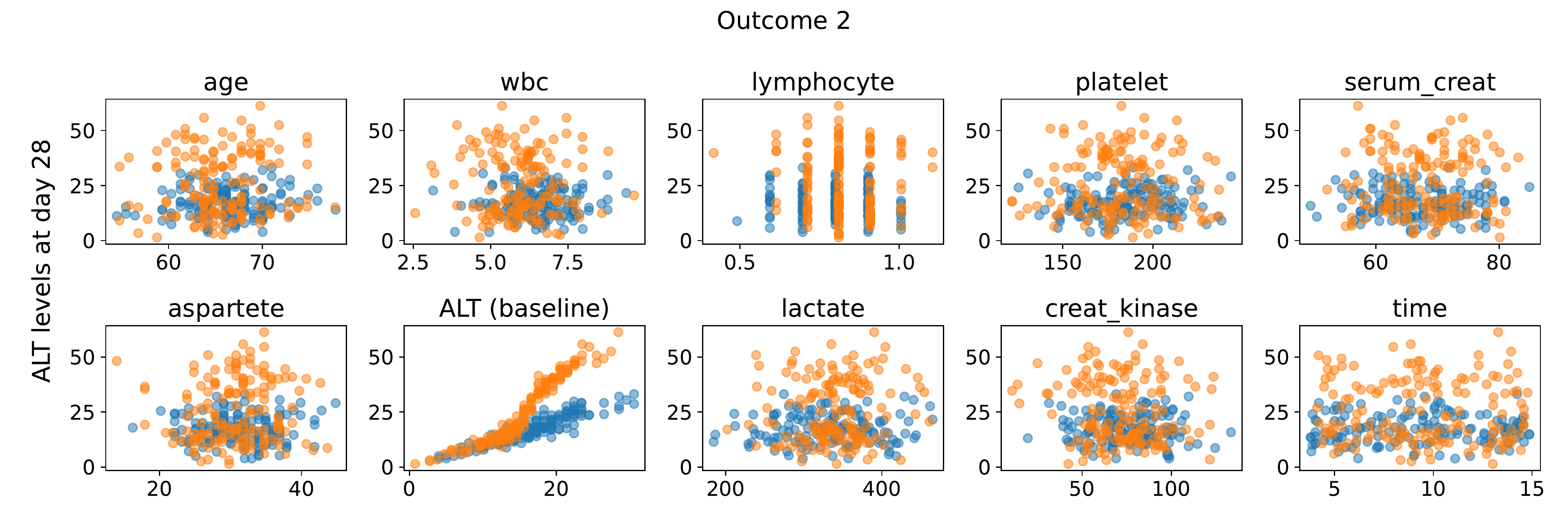}
\caption{\textbf{Treatment effect across covariates in the synthetic dataset, where the data is heteroscedastic.} Distributions are split across control and treatment populations, shown in blue and orange respectively.}
\label{fig:dist_synthb_hetsked}
\vspace{-1em} 
\end{figure*}

The treatment effect trends for the synthetic data across each covariate are shown in Figure \ref{fig:dist_synthb_uncorrout} (Figure \ref{fig:dist_synthb_corrcov} for correlated covariates, and Figure \ref{fig:dist_synthb_hetsked} for heteroscedastic data). The x-axis shows the values across each respective covariate and the y-axis shows the value of each respective outcome. Note the correlation between time and ALT level in each outcome.

\subsection{Semi-synthetic Data}

The first outcome showed the cognitive development score assessed by the Stanford-Binet Intelligence Scale, where infants enrolled in the intervention showed higher mean scores than the infants in the control population. Score differences were found to be dependent on the nnhealth of the infant. The second outcome evaluated the health status score of the infant measured by the mothers' report on the morbidity index. The health score showed positive treatment effect, but was dependent on the mother's age where younger mothers tended to report more frequent adverse health conditions. As part of the Response Surface B, other covariates in the dataset are randomly assign to have a small effect of the outcomes. 

\begin{figure*}[ht!]
\vspace{-5em} 
\centering
\includegraphics[scale=.39,trim={0cm 0cm 0cm 0cm}]{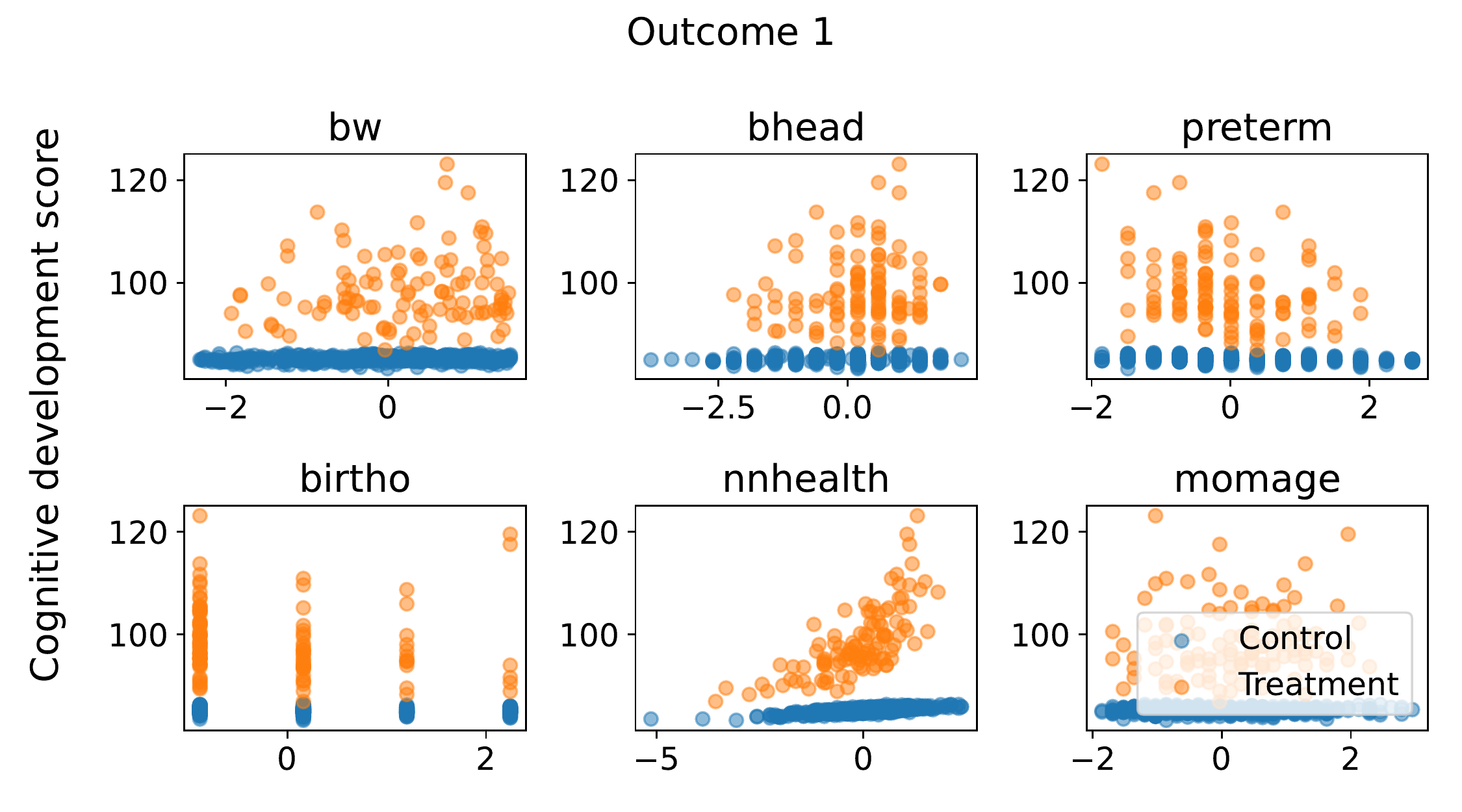}
\includegraphics[scale=.39,trim={0cm 0cm 0cm 0cm}]{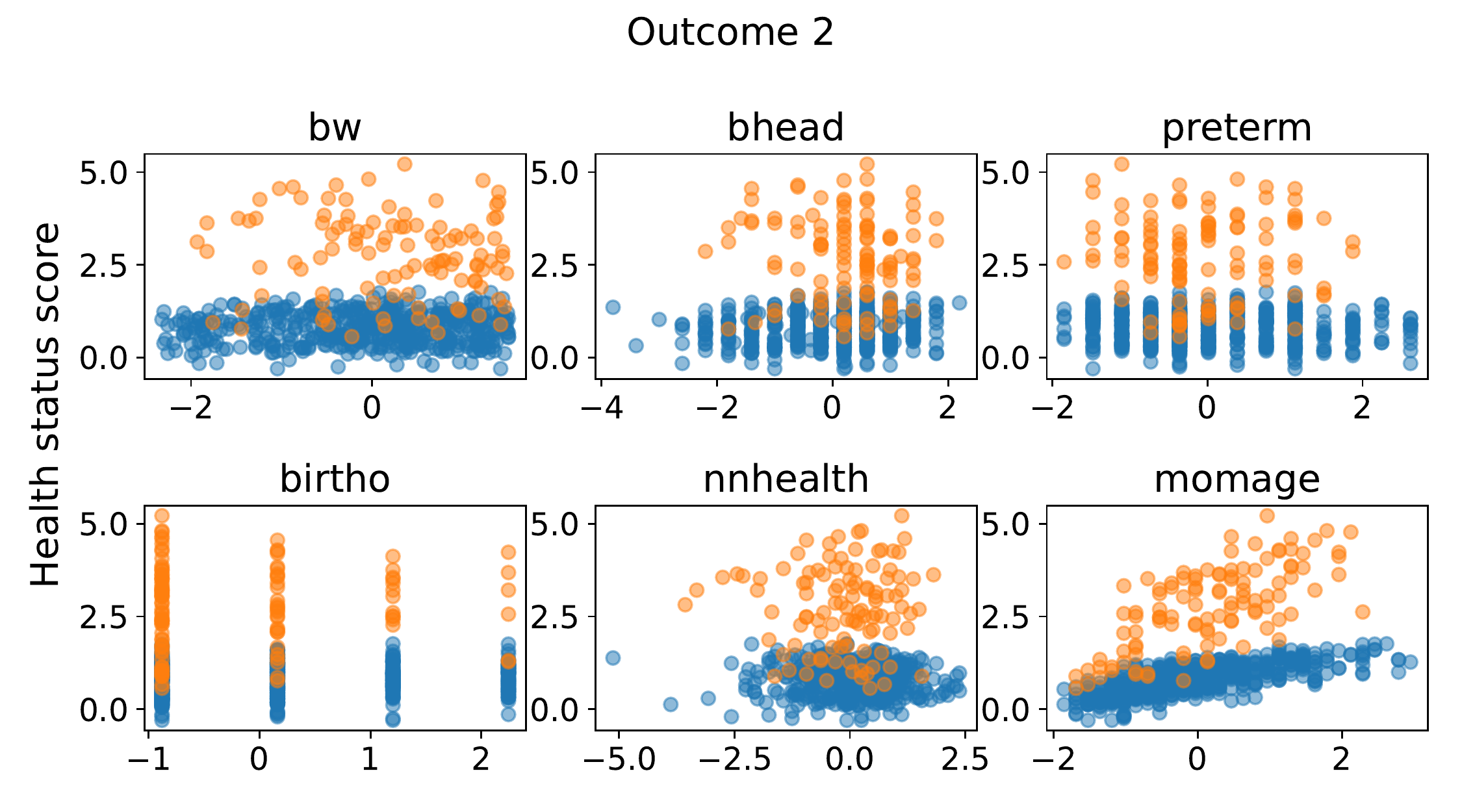}
\caption{\textbf{Treatment effect across covariates in the semi-synthetic dataset.} Distributions are split across control and treatment populations, shown in blue and orange respectively. Distributions are only shown in the 6 continuous covariates (remaining covariates are binary). Note. the continuous covariate values are normalized.}
\label{fig:dist_ihdp}
\end{figure*}

Treatment effects for the semi-synthetic data across each covariate are shown in Figure \ref{fig:dist_ihdp}. The x-axis details the values across each respective covariate and the y-axis shows the value of each respective outcome. Note the relationship between nnhealth and Outcome 1, and momage and Outcome 2. The dataset consisted of 608 untreated and 139 treated subjects, where the training and testing sets were split by 80\% and 20\%, respectively. The dataset included 25 covariates.

\newpage
\onecolumn
\twocolumn

\section{ITE Estimators}\label{app:ite}

The ITE estimators, Random Forest (RF) and Quantile Random Forest (QRF), were adapted from their original implementation in scikit-learn \citep{scikit} and scikit-garden \citep{scikit-garden} respectively. The adaptations were made in order to work in a conformal prediction framework. The results of the hyperparameter tuning for the RF on the synthetic data are found in Table \ref{tab:hyperparam_ite_synthB}; the hyperparameters for the semi-synthetic data are found in Table \ref{tab:hyperparam_ite_ihdp}. We chose to not tune the QRF, as the default parameters already gave high precision estimates.

\begin{table}[htbp!]
\centering
\begin{tabular}{ |l|l| } 
 \hline
 n\_estimators & 450 \\[0.1cm]
 random\_state & 0  \\[0.1cm]
 min\_samples\_split & 2 \\[0.1cm]
 min\_samples\_leaf & 1 \\[0.1cm]
 max\_depth & 38 \\[0.1cm]
 max\_features & auto \\[0.1cm]
 bootstrap & True \\[0.1cm]
 \hline
\end{tabular}
\caption{\label{tab:hyperparam_ite_synthB}Hyperparameters used in RF ITE estimator for the synthetic data.}
\end{table}

\begin{table}[htbp]
\centering
\begin{tabular}{ |l|l| } 
 \hline
 n\_estimators & 450 \\[0.1cm]
 random\_state & None  \\[0.1cm]
 min\_samples\_split & 3 \\[0.1cm]
 min\_samples\_leaf & 1 \\[0.1cm]
 max\_depth & 50 \\[0.1cm]
 max\_features & sqrt \\[0.1cm]
 bootstrap & False \\[0.1cm]
 \hline
\end{tabular}
\caption{\label{tab:hyperparam_ite_ihdp}Hyperparameters used in RF ITE estimator for the semi-synthetic data.}
\end{table}

\newpage
\onecolumn
\twocolumn

\section{Methodology}\label{app:methods}
\subsection{Assumptions}
The assumptions on which our methodology is based are as follows:
\begin{enumerate}
    \item This methodology holds in a RCT environment, where all patients have the same set of covariates and outcomes and there is no missingness.
    \item The outcomes are continuous values.
    \item There is an expected heterogeneous behavior in the dataset (for example in the medical space, the behavior may be proved through clinical references).
    \item Our methodology can be tested on multiple outcomes $\geq 2$, though for the purposes of this study, we focus on 2 outcomes.
\end{enumerate}

\subsection{Extension to Multiple Outcomes}

We provide an alternative formulation of the objective function below that can work in the setting of $d$ outcomes, where $\tau_{i}$ is the weight for outcome $i$. The equations for the objective functions for SCR and SCQR are below.
\begin{equation*}
\begin{split}
    \textrm{SCR: minimize} \sum_{g \in \Pi} \lambda \left(\sum_{i \in d} (\tau_{i} W^i_g)\right) + \\ (1 - \lambda) \left(\sum_{i \in d} (\tau_{i} V^i_g)\right) 
\end{split}
\end{equation*}
\begin{equation*}
\textrm{SCQR: minimize} \sum_{g \in \Pi}  \sum_{i \in d} (\tau_{i} V^i_g) 
\end{equation*}
\vspace{-2em}

\subsection{Partitioning Algorithm for SCR}
The partitioning algorithm for SCR can be found in Algorithm \ref{algo_partition_scr}.

\begin{algorithm2e}[tb]
\caption{SCR Recursive Partitioning on Two Outcomes}
  \KwIn{$G_{node}$, data} 
    \For{Covariate $j$ in data}{
        \For{Unique value $x$ of covariate $j$}{
            Split data in two branches on $x$ \\
            Compute $V^A$ and $W^A$ for each branch \\
            Compute $V^B$ and $W^B$ for each branch \\
            Set $ G_{split} $ to $G_{node}$ less the sum of $ \lambda (\beta W^A+(1-\beta)W^B) + (1 - \lambda)(\beta V^A+(1-\beta)V^B)$ from each branch\\
        }
    }
    Save best $G_{split}$ with covariate $j$ \\
    \uIf{$ G_{split}> \gamma G_{node}$ }{
        $ G_{node} \leftarrow G_{split}$ \\
        Partition on each branch \\
    }
    \Else{
        Set current node to leaf\\
    }
  \label{algo_partition_scr}
\end{algorithm2e}

\subsection{Hyperparameters}
\begin{figure*}[htpb!]
\centering
\includegraphics[scale=.3,trim={0cm 0cm 0cm 0cm}]{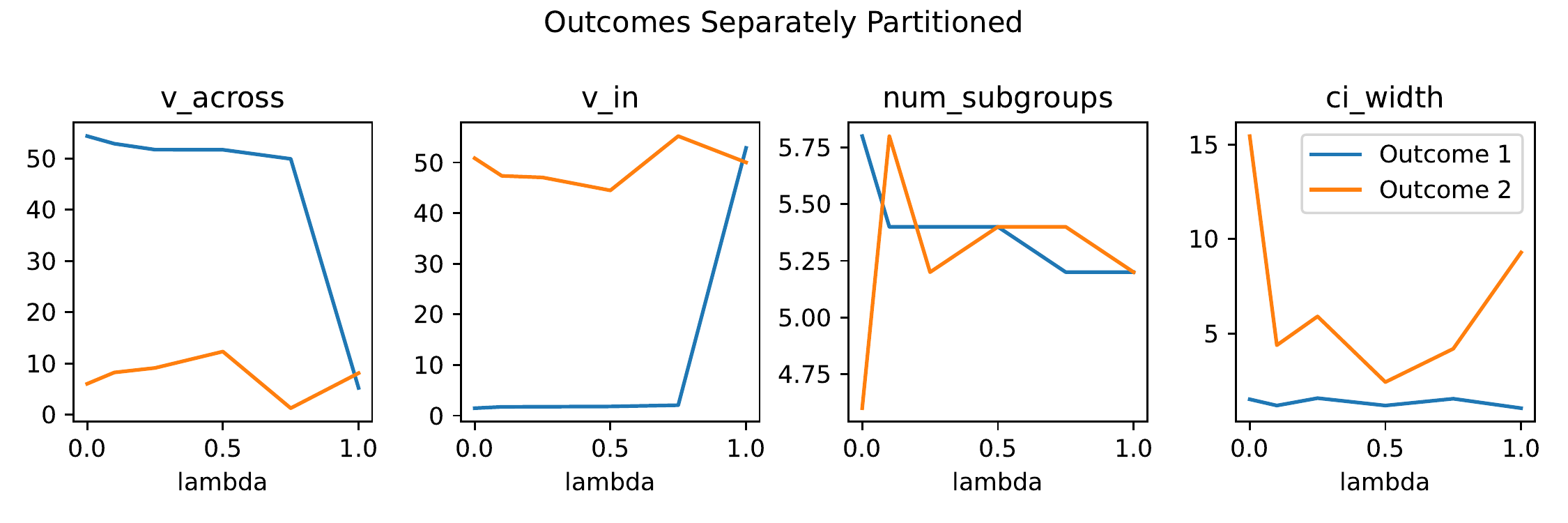}
\includegraphics[scale=.3,trim={0cm 0cm 0cm 0cm}]{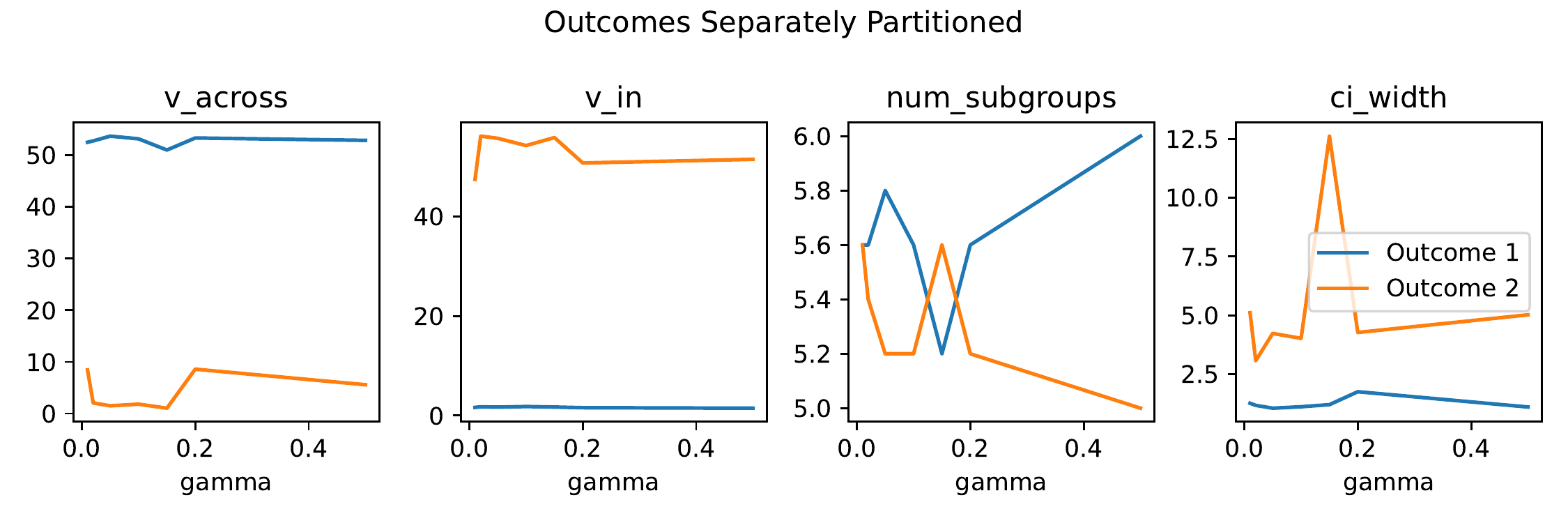}
\includegraphics[scale=.3,trim={0cm 0cm 0cm 0cm}]{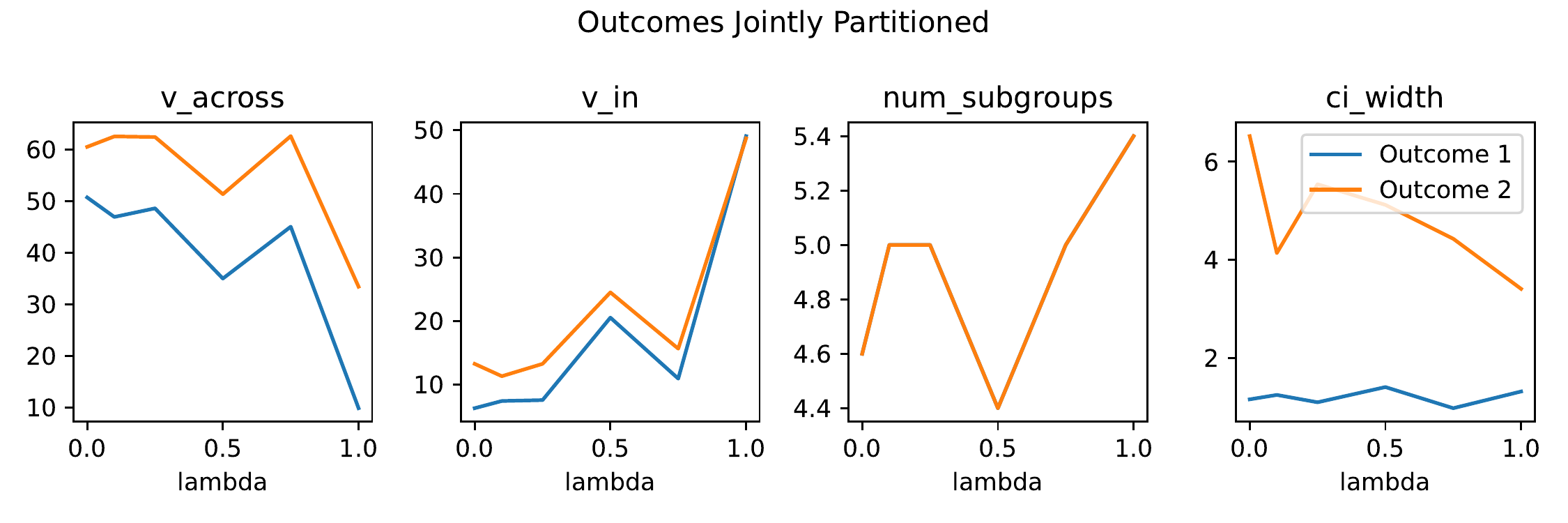}
\includegraphics[scale=.3,trim={0cm 0cm 0cm 0cm}]{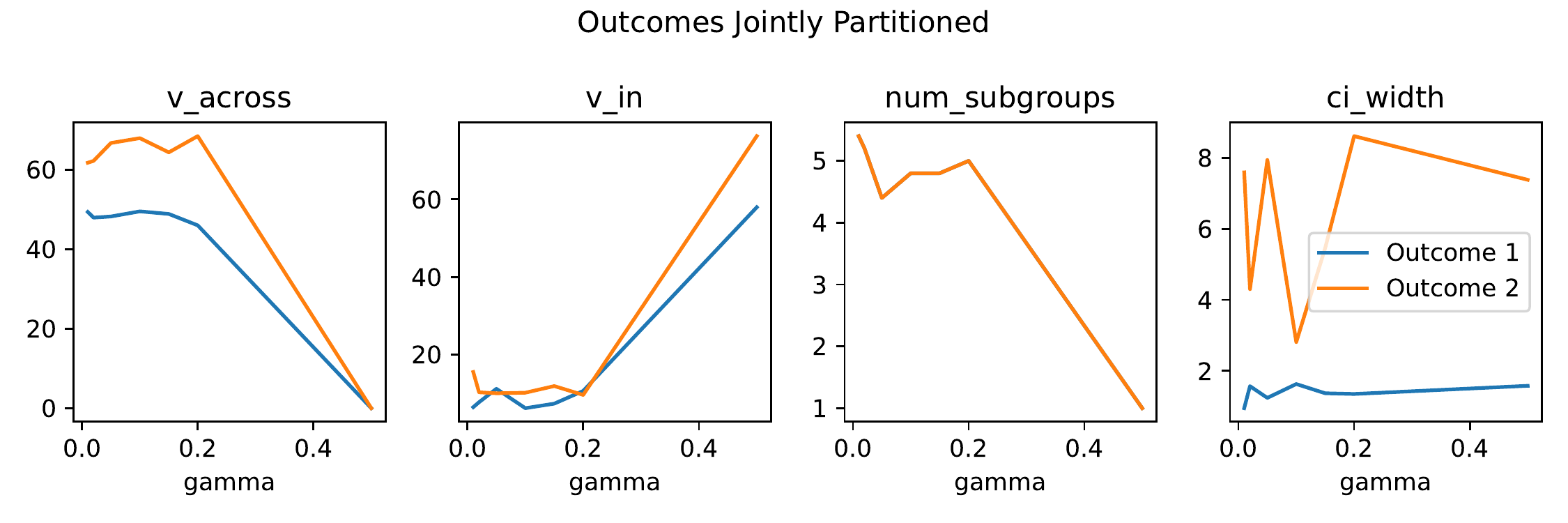}
\caption{Performance across separately and jointly partitioned outcomes varying $\lambda$ and $\gamma$ hyperparameters in the synthetic dataset.}
\label{fig:synth_hyperparam_tuning}
\end{figure*}

\begin{figure*}[htpb!]
\centering
\includegraphics[scale=.3,trim={0cm 0cm 0cm 0cm}]{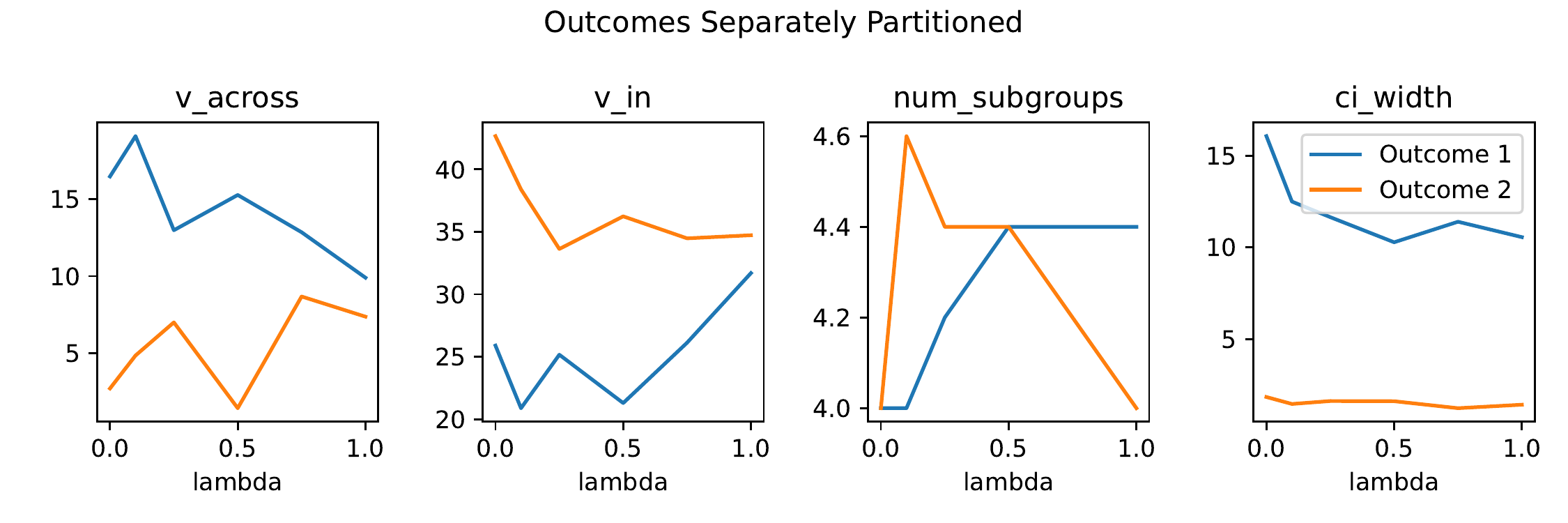}
\includegraphics[scale=.3,trim={0cm 0cm 0cm 0cm}]{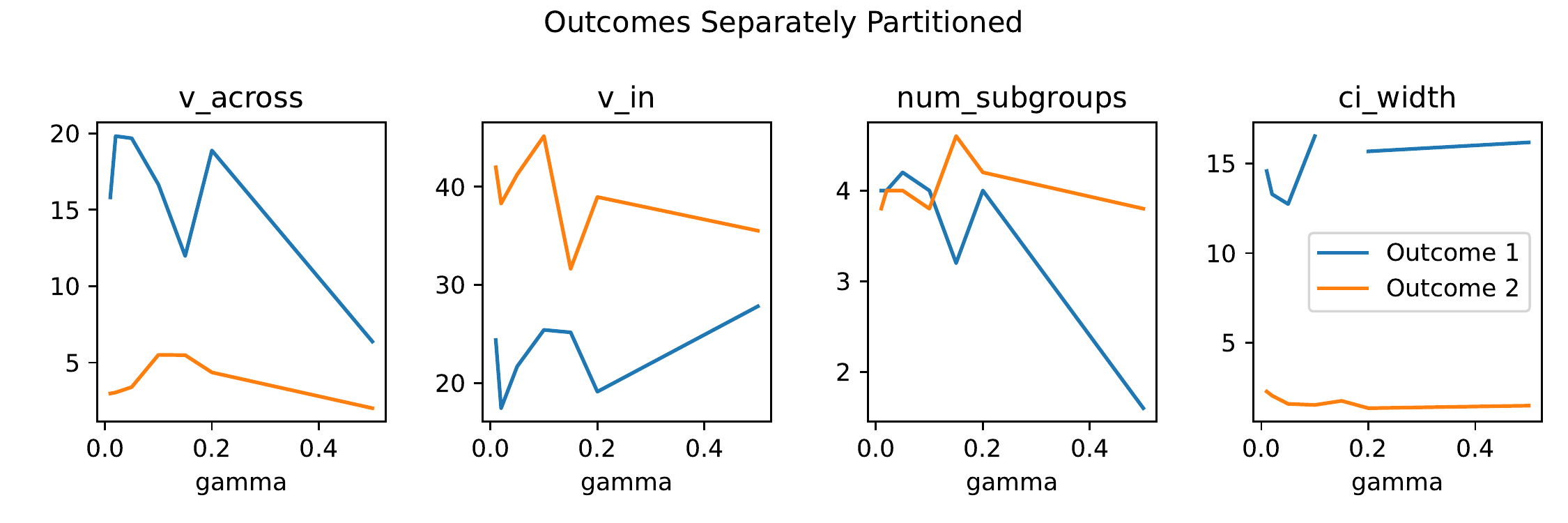}
\includegraphics[scale=.3,trim={0cm 0cm 0cm 0cm}]{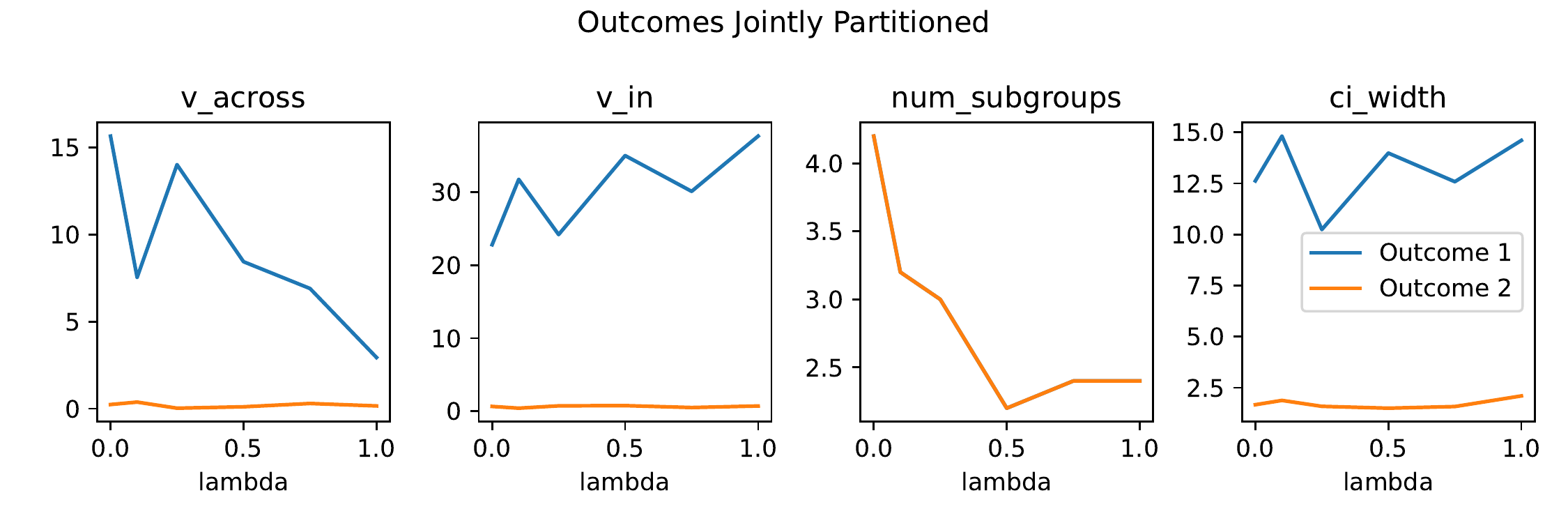}
\includegraphics[scale=.3,trim={0cm 0cm 0cm 0cm}]{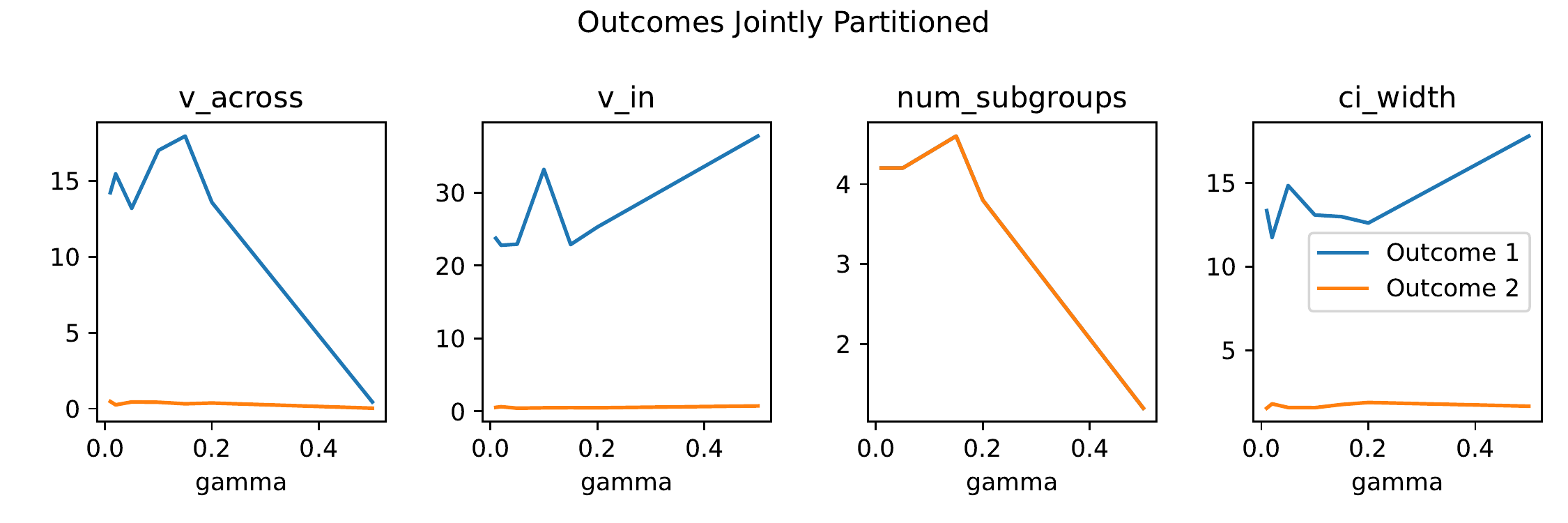}
\caption{Performance across separately and jointly partitioned outcomes varying $\lambda$ and $\gamma$ and hyperparameters in the semi-synthetic dataset.}
\label{fig:semisynth_hyperparam_tuning}
\setlength{\belowcaptionskip}{-10pt}
\end{figure*}

Hyperparameter tuning was conducted on the ITE estimators using random forests as shown in Appendix \ref{app:ite}. As for the hyperparameters in the partitioning algorithm, $\lambda$ and $\gamma$ were tested across varying values and set to the values shown in Table \ref{tab:hyperparam_synth} (such that overall, V\_across was maximized, V\_in was minimized and ci\_width was minimized). $\lambda$ is used to vary the weight between the expected absolute deviation within a group and the CI width, affecting the number of subgroups and the inter- and intra-subgroup variance. $\gamma$ controls for regularization, where too small of a value can lead to overfitting with a large number of subgroups and too large of a value can lead to poor performance with a small number of subgroups. The effects of varying $\lambda$,and $\gamma$ in the synthetic dataset is shown in Figure \ref{fig:synth_hyperparam_tuning} (tuning for the semi-synthetic dataset is shown in Figure \ref{fig:semisynth_hyperparam_tuning}).

\begin{figure*}[htpb!]
\centering
\includegraphics[scale=.3,trim={0cm 0cm 0cm 0cm}]{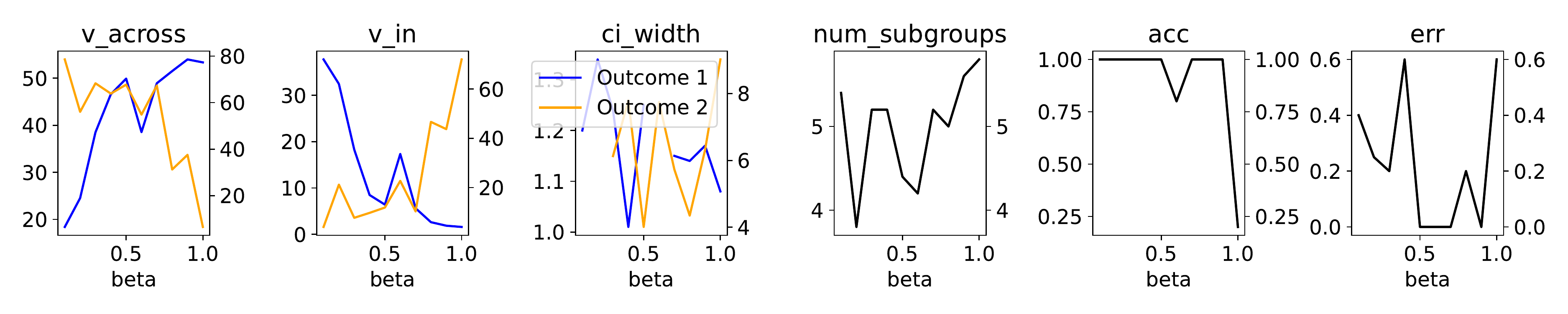}
\caption{$\beta$ tuning using SCR CMGP on the synthetic data}
\label{fig:beta synthetic1}
\end{figure*}

\begin{figure*}[htpb!]
\centering
\includegraphics[scale=.3,trim={0cm 0cm 0cm 0cm},clip]{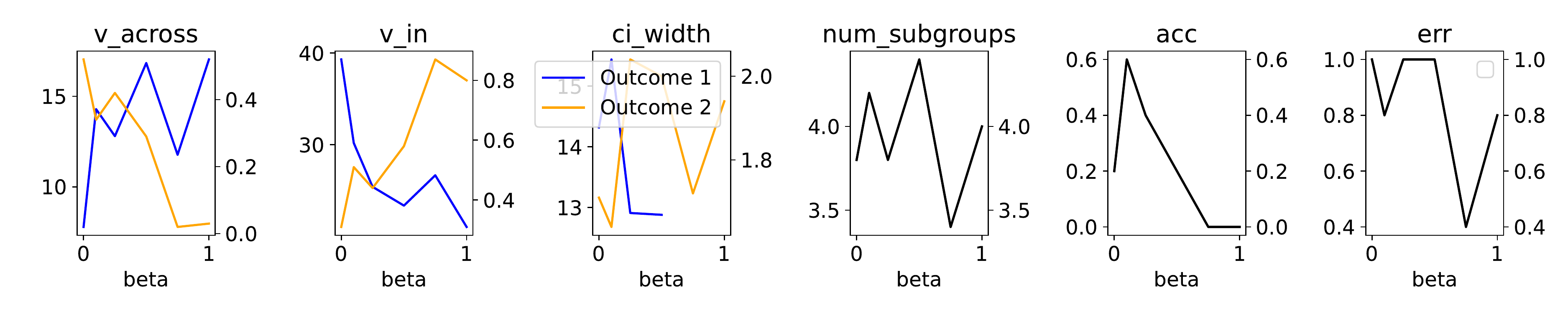}
\caption{$\beta$ tuning using SCR CMGP on the semi-synthetic data}
\label{fig:beta semi synthetic}
\end{figure*}

In our experiments, $\beta$ determines the weight of each outcome in the algorithm. In the two outcome case that we have explored, a $\beta$ value other than .5 will favor one outcome over the other. This parameter can be used to prioritize partitioning on one outcome more than the other. Additionally, in scenarios where the magnitudes of the treatment effects of each outcome are very different, $\beta$ can be used to weight the effect accordingly. We tested the impact that $\beta$ has on certain metrics for both datasets. Figure \ref{fig:beta synthetic1} shows the effect of $\beta$ on the performance metrics for the synthetic dataset. Figure \ref{fig:beta semi synthetic} shows the effect in the IHDP dataset. Since in the IHDP dataset, outcome 1 has a higher magnitude than outcome 2, setting $\beta$ to be a lower than .5 allows the algorithm to find heterogeneity in outcome 2. Since many covariates contribute to both outcomes in IHDP, the error metric is not reported.

To generate the confidence regions $W$ and $V$, we use the miscoverage rates of .1 and .8, respectively. 

\begin{table}[htbp]
\centering
\begin{tabular}{ |c|c|c| } 
 \hline
 & Synthetic & Semi-synthetic \\\hline
 $\lambda$ & 0.25 & 0 \\[0.1cm]
 $\gamma$ & 0.05 & 0.02 \\[0.1cm]
 $\beta$ & 0.5 & 0.25 \\[0.1cm]
 \hline
\end{tabular}
\caption{\label{tab:hyperparam_synth}Hyperparameters used in the partitioning algorithm for each dataset.}
\vspace{-2em}
\end{table}

\subsection{Licenses}

The license of the assets used in this paper are as follows:
\begin{itemize}
    \item Robust recursive partitioning algorithm, and synthetic data:\url{https://github.com/vanderschaarlab/mlforhealthlabpub/blob/main/LICENSE.md}
    \item SCR:\url{https://github.com/ryantibs/conformal/blob/master/LICENSE}
    \item SCQR:\url{https://github.com/yromano/cqr/blob/master/LICENSE}
    \item IHDP dataset: A license was not provided, the code was zipped in the supplementary material of \citep{Hill2011}.
\end{itemize}

\newpage
\onecolumn
\twocolumn

\section{Additional Results}\label{app:results}

In this section we show additional results that were omitted from the paper. We show the tables and figures for the results on the two versions from the synthetic data that were not discussed in the paper. These are correlated covariates and heteroscedastic data. We additionally show subgroup analyses for all results on the synthetic and semi-synthetic data. In these subgroup analyses we show a sample subgroup from each method and the characteristics of each subgroup. 

\subsection{Separate Versus Joint Partitioning} 
\label{app:sep_vs_joint_res}
\begin{figure*}[ht!]
  \vspace{-2.5em}
  \centering
  \includegraphics[scale=.3,trim={0cm 0cm 0cm 0cm}]{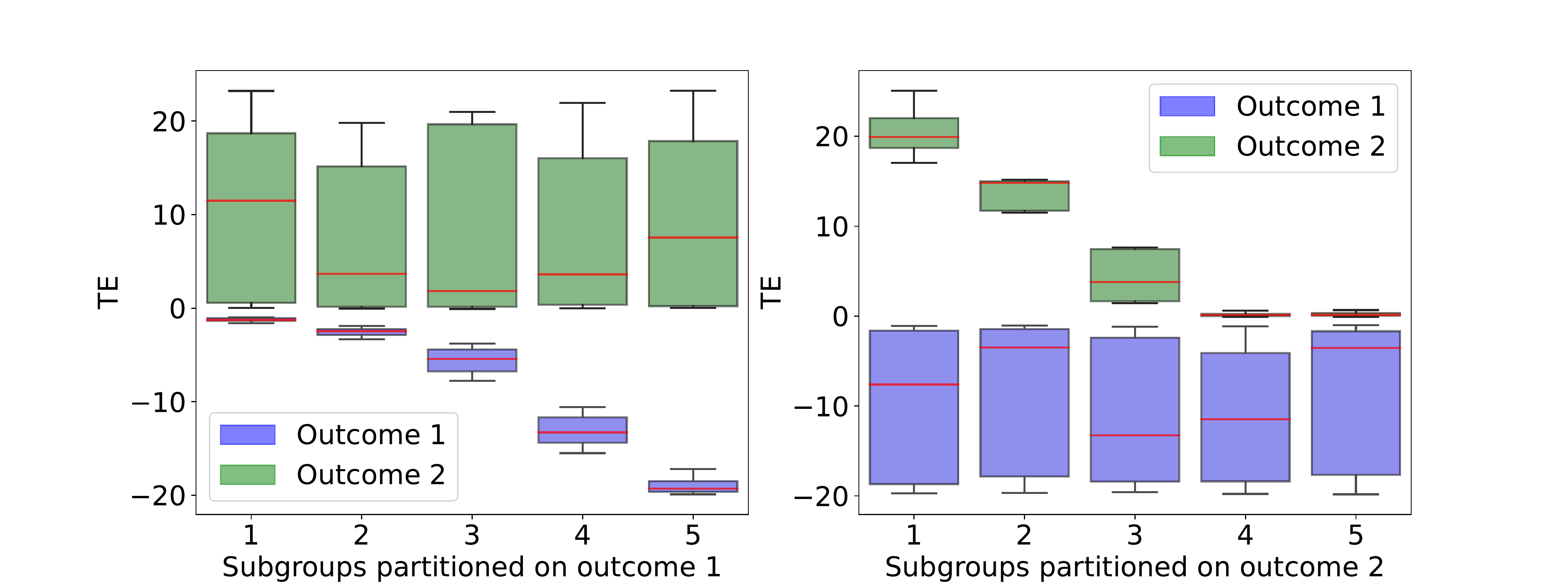}
  \caption{\textbf{Separately partitioned subgroups on synthetic data using the CMGP estimator (Baseline R2P).} Subgroups defined in each outcome when partitioned on a single treatment outcome individually. Left plot shows the box plot for treatment effects when the covariates are partitioned using outcome 1 only. The right plot shows the box plots when the covariates are partitioned using outcome 2 only. Whiskers show the 25th and 75th percentiles.}
  \label{fig:uncorr_scr_cmgp_sep}
  \vspace{-2.5em}
\end{figure*}

\begin{figure*}[ht!]
  \centering
  \includegraphics[scale=.3,trim={0cm 0cm 0cm 0cm}]{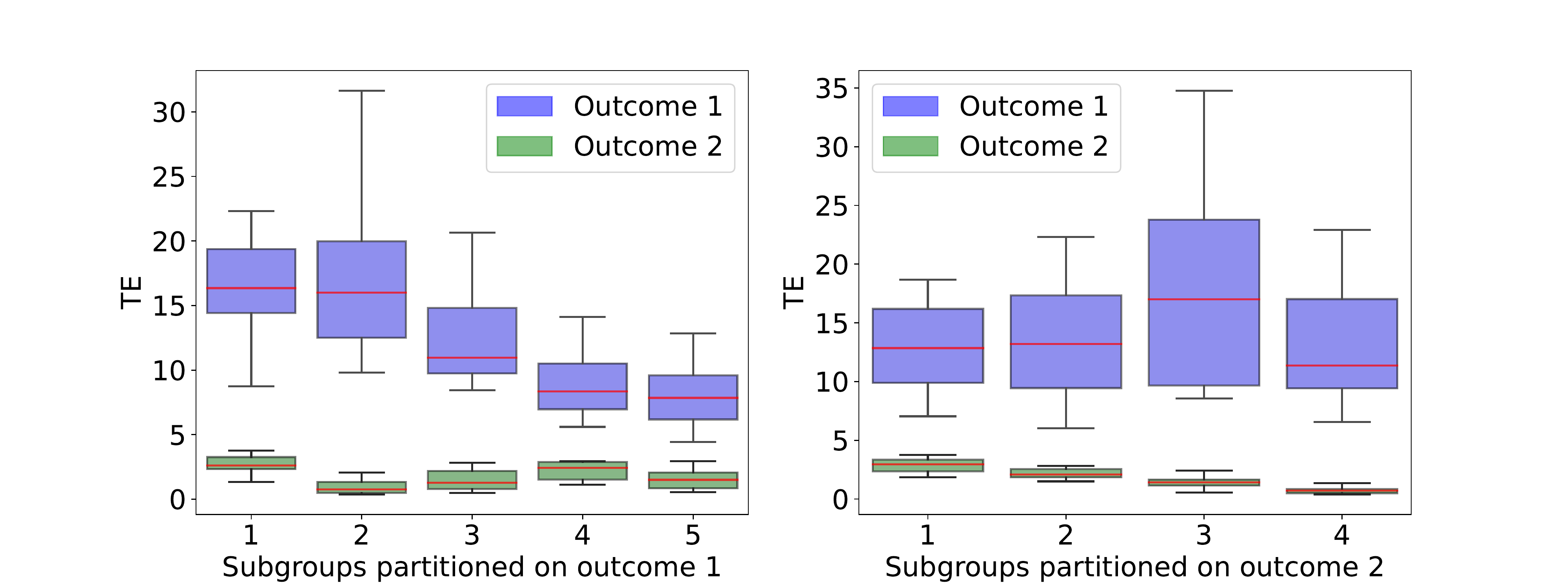}
  \caption{\textbf{Separately partitioned subgroups on semi-synthetic data using the CMGP estimator (Baseline R2P).} Subgroups defined in each outcome when partitioned on a single treatment outcome individually. Left plot shows the box plot for treatment effects when the covariates are partitioned using outcome 1 only. The right plot shows the box plots when the covariates are partitioned using outcome 2 only. Whiskers show the 25th and 75th percentiles.}
  \label{fig:ihdp_1_scr_cmgp_sep}
  \vspace{-2.5em}
\end{figure*}

\begin{figure*}[ht!]
  \centering 
  (a){\includegraphics[scale=.3,trim={0cm 0cm 0cm 0cm}]{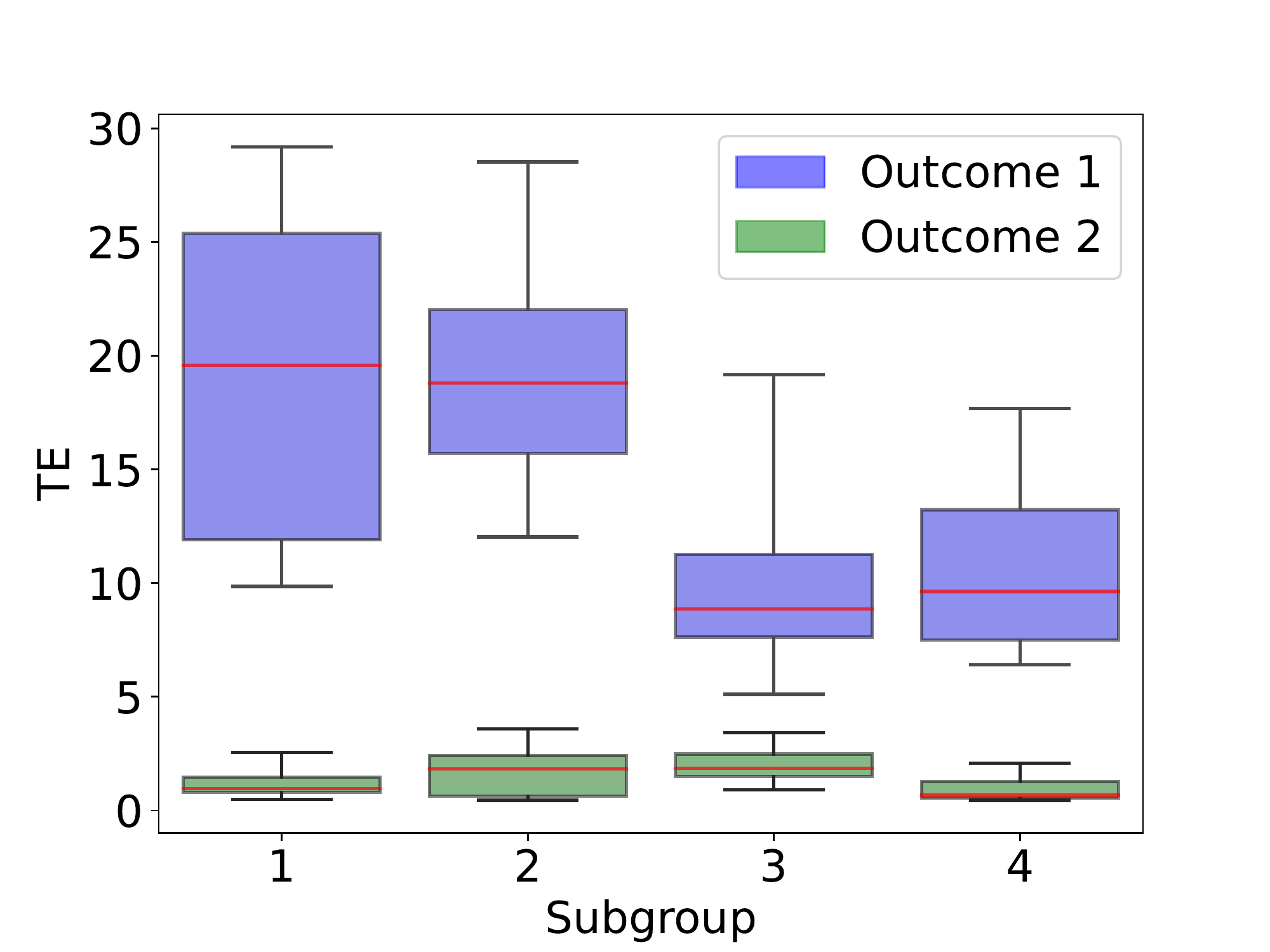}}
  (b){\includegraphics[scale=.3,trim={0cm 0cm 0cm 0cm}]{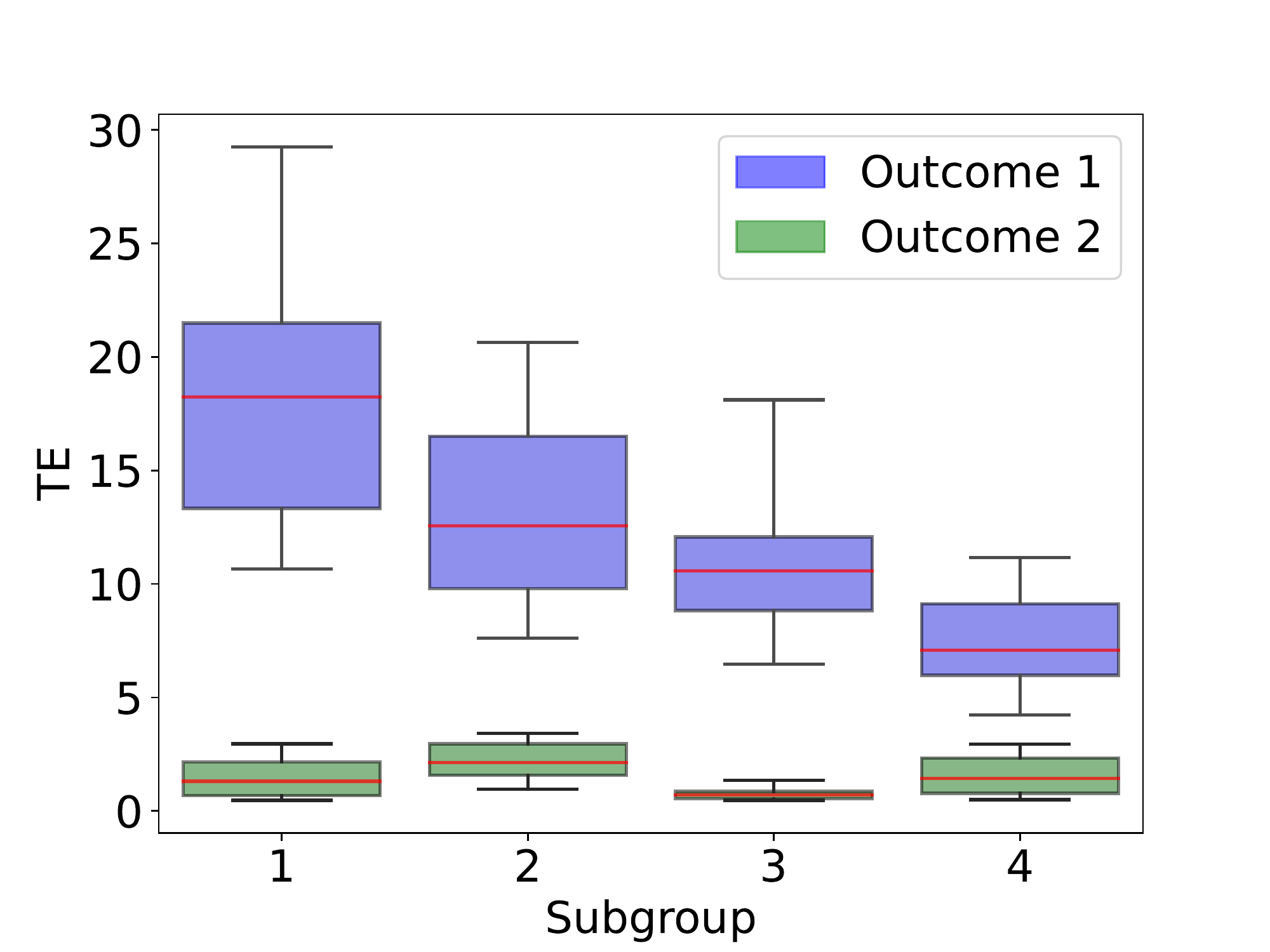}}
  \caption{\textbf{Jointly partitioned subgroups on semi-synthetic data (MOP-JCI).} Subgroups defined in each outcome when partitioned using the RF estimator on the SCR method (a) and the quantile method (b). Whiskers show the 25th and 75th percentiles.}
  \label{fig:ihdp_1_joint}
  \vspace{-2em}
\end{figure*}

Separately partitioned subgroups using the CMGP estimator (Baseline R2P) are shown on synthetic data (Figure \ref{fig:uncorr_scr_cmgp_sep}) and on semi-synthetic data (Figure \ref{fig:ihdp_1_scr_cmgp_sep}). Jointly partitioned subgroups on semi-synthetic data using our method (MOP-JCI) are shown in Figure \ref{fig:ihdp_1_joint}.
\vspace{-0.3em}

\subsection{Correlated Covariates}
\begin{figure*}[ht!]
  \vspace{-2.5em}
  \centering
  \includegraphics[scale=.3,trim={0cm 0cm 0cm 0cm}]{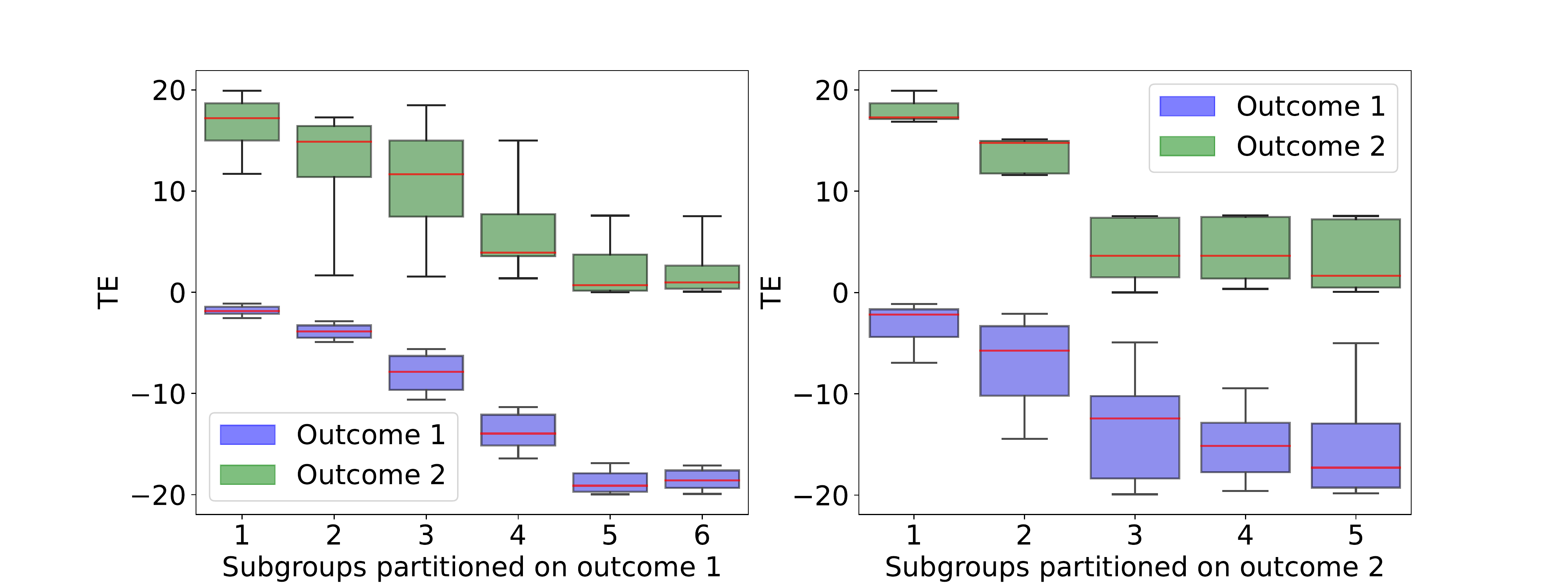}
  \caption{\textbf{Separately partitioned subgroups on synthetic data with correlated covariates using the CMGP estimator (Baseline R2P).} Subgroups defined in each outcome when partitioned on a single treatment outcome individually. Left plot shows the box plot for treatment effects when the covariates are partitioned using outcome 1 only. The right plot shows the box plots when the covariates are partitioned using outcome 2 only. Whiskers show the 25th and 75th percentiles.}
  \label{fig:corr_cov_scr_cmgp_sep}
  \vspace{-2.5em}
\end{figure*}

\begin{figure*}[ht!]
  \centering 
  (a){\includegraphics[scale=.3,trim={0cm 0cm 0cm 0cm}]{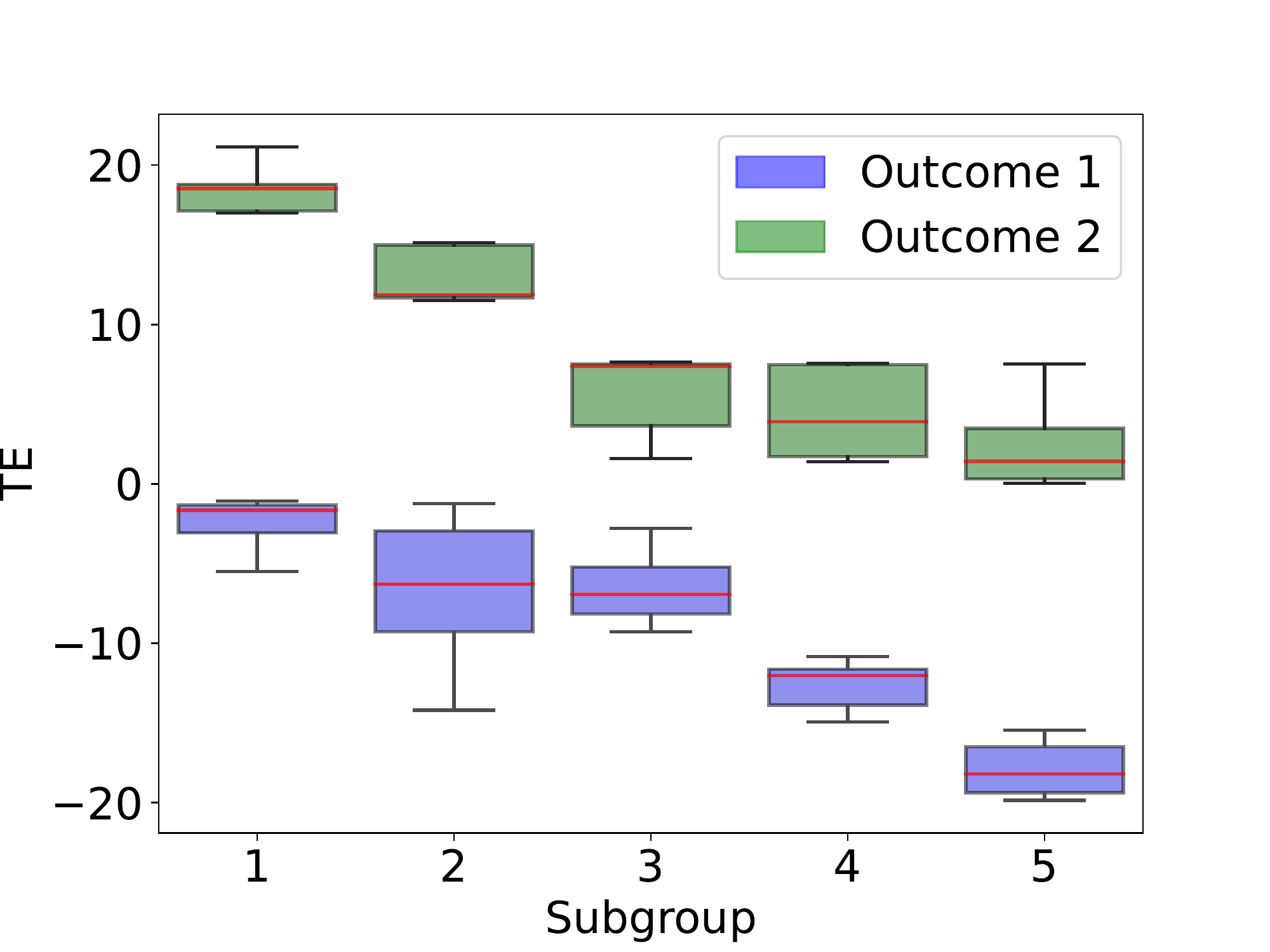}}
  (b){\includegraphics[scale=.3,trim={0cm 0cm 0cm 0cm}]{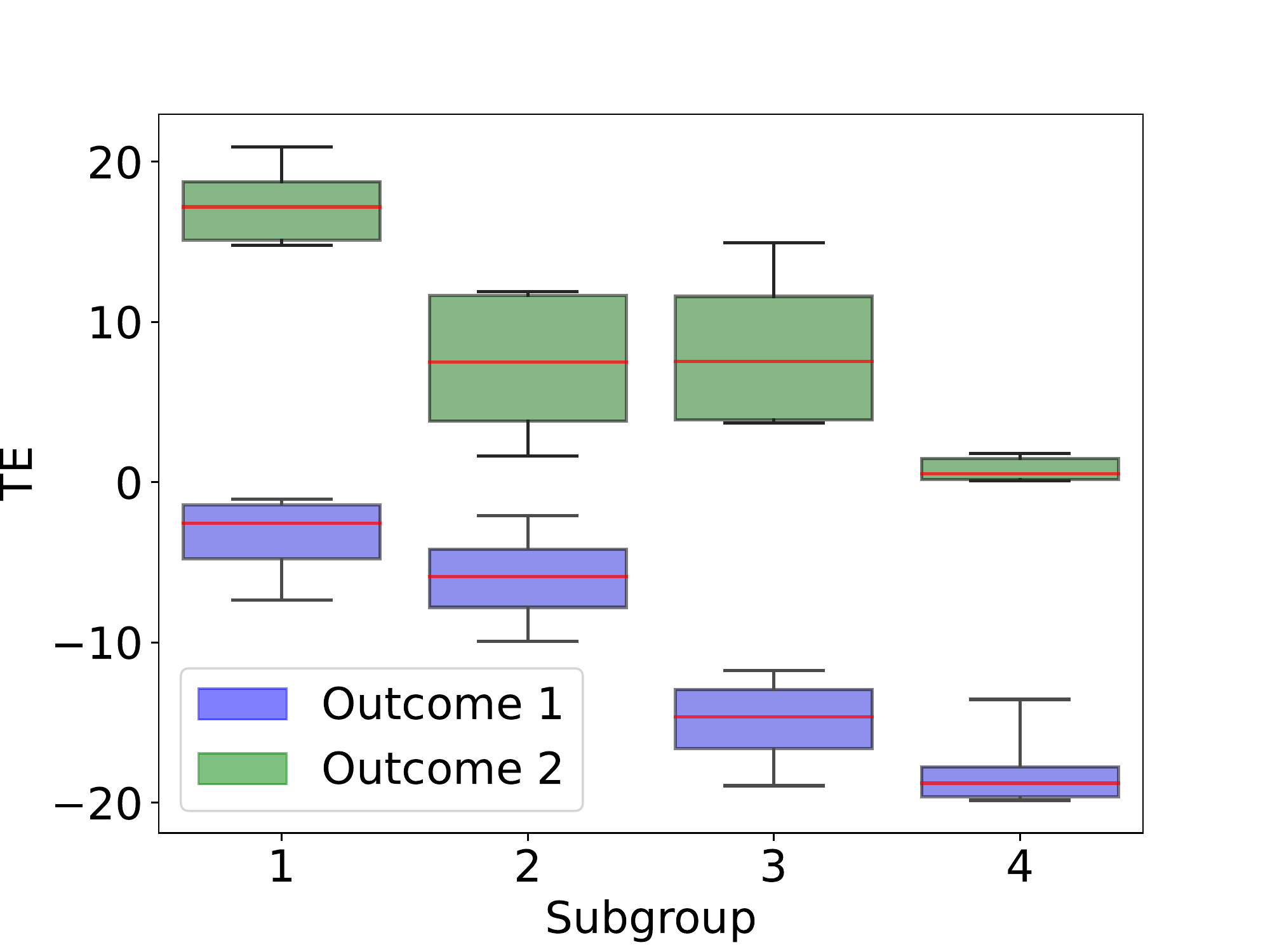}}
  \caption{\textbf{Jointly partitioned subgroups on synthetic data with correlated covariates (MOP-JCI).} Subgroups defined in each outcome when partitioned using the RF estimator on the SCR method (left) and the quantile method (right). Whiskers show the 25th and 75th percentiles.}
  \label{fig:corr_cov_joint}
\end{figure*}

Here, we show the results from the synthetic data with correlated covariates. Figure \ref{fig:corr_cov_scr_cmgp_sep} shows the treatment effect for both outcomes of each subgroup when partitioned on each outcome separately. We show the CMGP ITE estimator method. Figure \ref{fig:corr_cov_joint} shows the treatment effect for both outcomes of each subgroup when partitioned on each outcome jointly. We show the method using SCR and SCQR, both using a RF estimator. 

In Table \ref{tab:synth_corr_cov}, we show the numerical results of each subgroup with mean and standard deviation when running each method 30 times. We show the variance across each group, the variance within each group, the precision of the ITE estimator, the coverage, and the average CI of each subgroup. The PEHE is computed using the 50th quantile for the SCQR method. 

\begin{table*}[t!]
\scriptsize
\centering
\setlength{\tabcolsep}{1pt} 
\begin{tabular}{p{3cm}|p{1cm}|p{1cm}p{1cm}p{1cm}p{1cm}p{1cm}|p{1cm}p{1cm}p{1cm}p{1cm}p{1cm}}
\toprule
\multicolumn{12}{l}{Baselines (R2P)} \\\hline
&  & \multicolumn{5}{c|}{Outcome 1} & \multicolumn{5}{c}{Outcome 2} \\\hline
&   Num groups &  $V_{across}$ &   $V_{within}$ &   PEHE &  CI Width &   Cov        &  $V_{across}$ &   $V_{within}$ &   PEHE &  CI Width  & Cov \\\hline
CMGP on outcome 1 &     5.17 ±0.24 &   \textbf{40.68 ±1.11} &     \textbf{1.57 ±0.13} &     0.41 ±0.11 &     1.94 ±0.28 &    97.95 ±0.59 &   23.92 ±1.14 &    18.09 ±0.83 &                - &                - &                - \\
CMGP on outcome 2 &     5.03 ±0.18 &   25.90 ±1.23 &    16.44 ±1.04 &                - &                - &                - &   \textbf{40.21 ±1.81} &     \textbf{2.43 ±0.59} &     0.35 ±0.09  &     2.56 ±0.64 &    98.62 ±0.41 \\
RF on outcome 1 &     5.03 ±0.25 &   40.11 ±1.51 &     1.70 ±0.12 &     0.51 ±0.03  &     3.77 ±0.34 &    99.45 ±0.35 &   24.36 ±1.58 &    18.11 ±0.77 &                - &                - &                - \\
RF on outcome 2 &     5.17 ±0.24 &   24.54 ±1.81 &    17.21 ±1.24 &                - &                - &                - &   37.79 ±1.74 &     4.02 ±0.76 &     0.66 ±0.07  &     6.45 ±0.56 &    99.08 ±0.55 \\
\bottomrule
\end{tabular}

\setlength{\tabcolsep}{1pt}
\begin{tabular}{p{2cm}|p{1cm}p{1cm}p{1cm}|p{1cm}p{1cm}p{1cm}p{1cm}|p{1cm}p{1cm}p{1cm}p{1cm}|p{1cm}}
\toprule
\multicolumn{12}{l|}{Jointly Partitioned (MOP-JCI)} \\\hline
&  & & & \multicolumn{4}{c|}{Outcome 1} & \multicolumn{4}{c|}{Outcome 2} & \\\hline
&     Num groups & Split Acc &          Split Err &       $V_{across}$ &        $V_{within}$ &            PEHE &               CI Width &       $V_{across}$ &        $V_{within}$ &            PEHE &               CI Width &     Cov (joint) \\ \hline
CMGP (SCR) &    5.00 ±0.22 &        97\% &        17\% &   35.88 ±1.58 &     5.35 ±0.76 &     0.48 ±0.19 &     1.80 ±0.31 &   \textbf{33.93 ±1.93} &     \textbf{7.91 ±1.29} &     0.39 ±0.08 &     2.67 ±0.80 &    97.13 ±0.74 \\
RF (SCR) &    5.03 ±0.25 &        97\% &        27\% &   \textbf{35.94 ±1.48} &     4.56 ±0.73 &     0.52 ±0.04 &     3.90 ±0.40 &   31.89 ±2.01 &     9.52 ±1.42 &     0.67 ±0.09 &     6.60 ±0.68 &    99.55 ±0.28 \\
QRF (SCQR) &    4.97 ±0.18 &        93\% &        13\% &   35.21 ±1.70 &     \textbf{4.01 ±0.60} &     0.60 ±0.04 &     5.63 ±0.41 &   30.62 ±1.94 &     9.95 ±1.63 &     0.72 ±0.07 &     9.27 ±0.82 &    97.55 ±1.65 \\
\bottomrule
\end{tabular}
\caption{Results from synthetic data with correlated covariates. We take the mean and standard deviation of each metric across 30 runs. Num groups is the number of subgroups generated. Best performance for $V_{across}$ and $V_{within}$ in each column are in bold. }
\label{tab:synth_corr_cov}
\end{table*}

\subsection{Heteroscedasticity}
\begin{figure*}[ht!]
  \vspace{-5em}
  \centering
  \includegraphics[scale=.3,trim={0cm 0cm 0cm 0cm}]{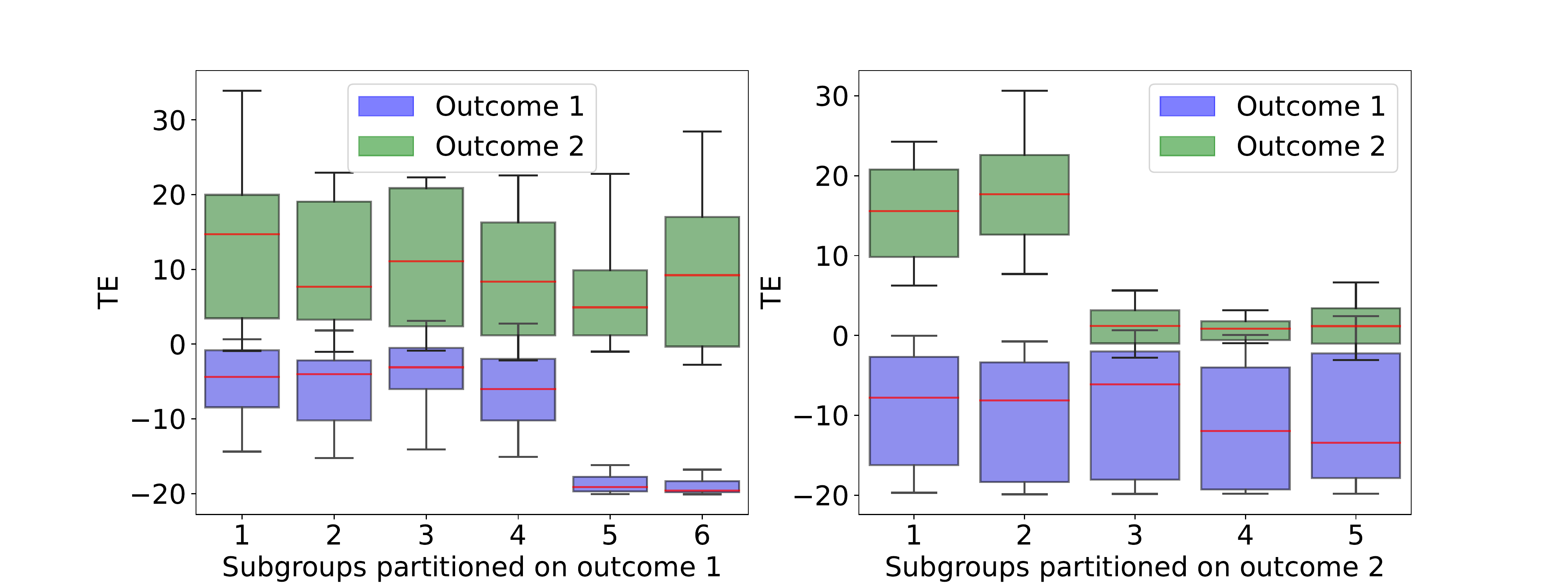}
  \caption{\textbf{Separately partitioned subgroups on heteroscedastic synthetic data using the CMGP estimator (Baseline R2P).} Subgroups defined in each outcome when partitioned on a single treatment outcome individually. Left plot shows the box plot for treatment effects when the covariates are partitioned using outcome 1 only. The right plot shows the box plots when the covariates are partitioned using outcome 2 only. Whiskers show the 25th and 75th percentiles.}
  \label{fig:hetsked_scr_cmgp_sep}
\end{figure*}

\begin{figure*}[ht!]
  \vspace{-5em}
  \centering 
  (a){\includegraphics[scale=.3,trim={0cm 0cm 0cm 0cm}]{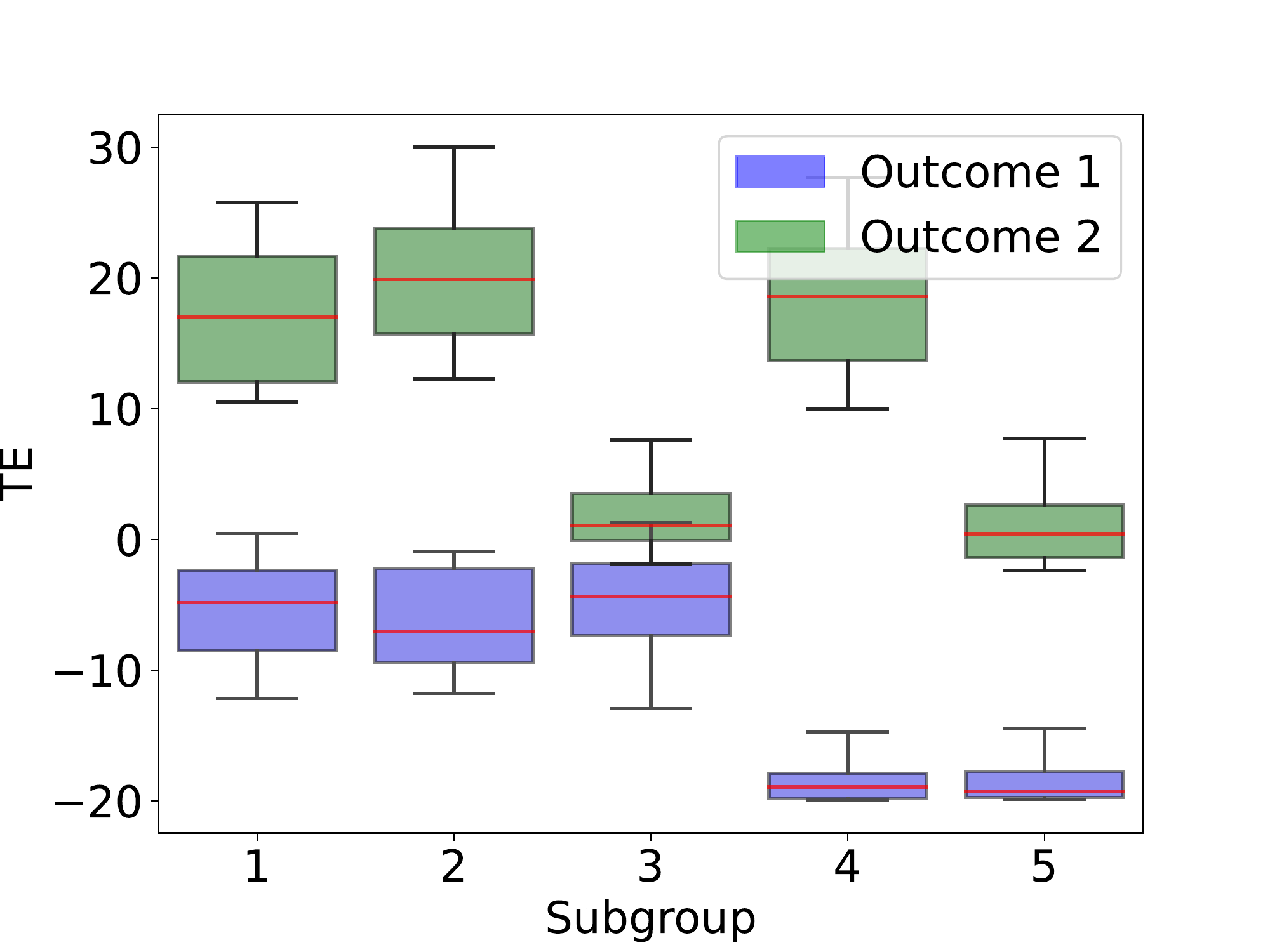}}
  (b){\includegraphics[scale=.3,trim={0cm 0cm 0cm 0cm}]{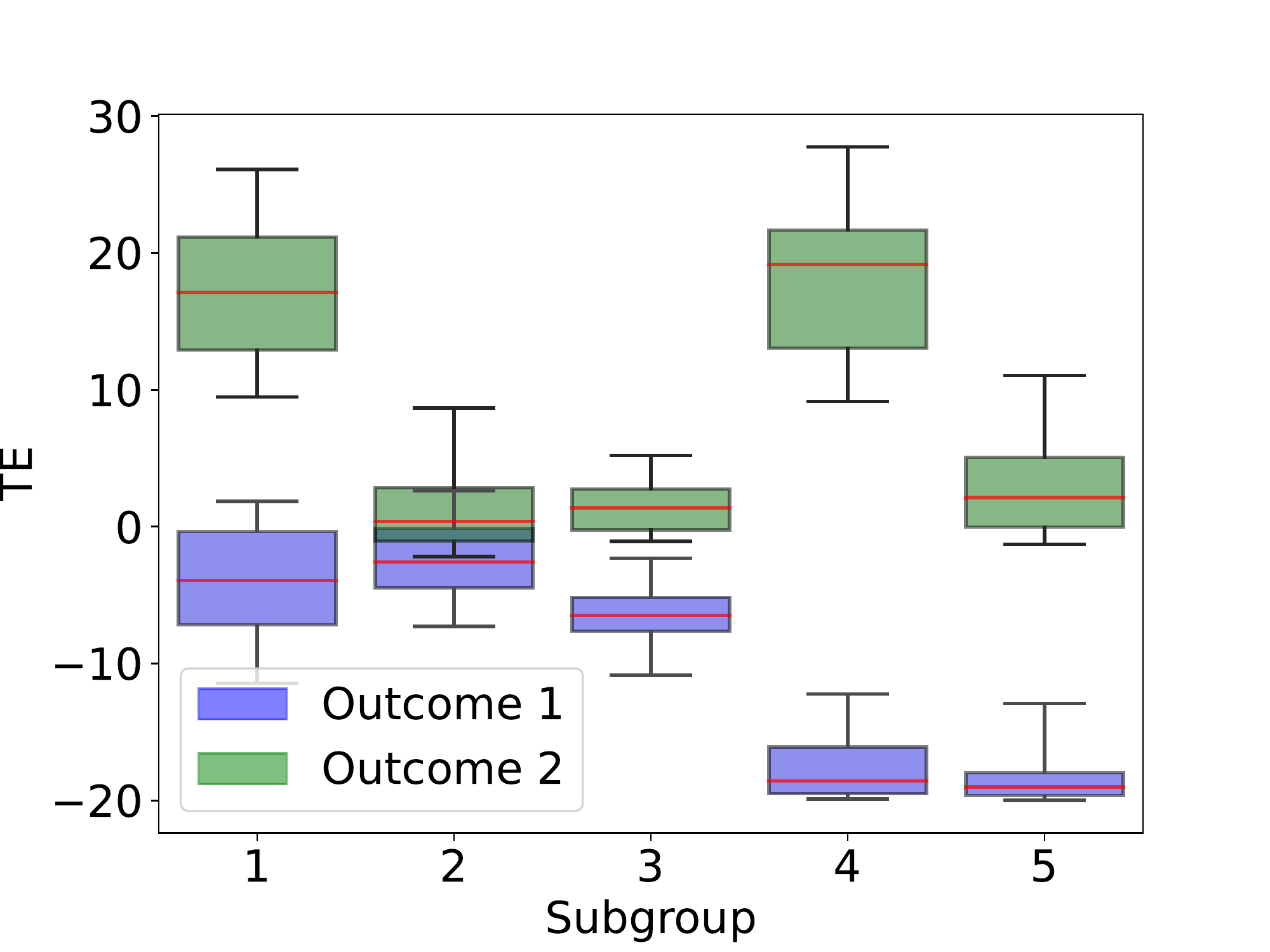}}
  \caption{\textbf{Jointly partitioned subgroups on heteroscedastic synthetic data (MOP-JCI).} Subgroups defined in each outcome when partitioned using the RF estimator on the SCR method (left) and the SCQR method (right). Whiskers show the 25th and 75th percentiles.}
  \label{fig:hetsked_joint}
\end{figure*}

Here, we show the results from the synthetic data with added heteroscedasticity. Figure \ref{fig:hetsked_scr_cmgp_sep} shows the treatment effect for both outcomes of each subgroup when partitioned on each outcome separately. We show the CMGP ITE estimator method. Figure \ref{fig:hetsked_joint} shows the treatment effect for both outcomes of each subgroup when partitioned on each outcome jointly. We show the method using SCR and SCQR, both using a RF estimator. 

In Table \ref{tab:synth_hetsked} we show the numerical results of each partitioning with mean and standard deviation when running each method 30 times. We show the variance across each group, the variance within each group, the precision of the ITE estimator, the coverage, and the average CI of each subgroup. The PEHE is computed using the 50th quantile for the SCQR method. 

\begin{table*}[htbp!]
\scriptsize
\centering
\setlength{\tabcolsep}{1pt} 
\begin{tabular}{p{3cm}|p{1cm}|p{1cm}p{1cm}p{1cm}p{1cm}p{1cm}|p{1cm}p{1cm}p{1cm}p{1cm}p{1cm}}
\toprule
\multicolumn{12}{l}{Baselines (R2P)} \\\hline
&  & \multicolumn{5}{c|}{Outcome 1} & \multicolumn{5}{c}{Outcome 2} \\\hline
&   Num groups &  $V_{across}$ &   $V_{within}$ &   PEHE &  CI Width &   Cov        &  $V_{across}$ &   $V_{within}$ &   PEHE &  CI Width  & Cov \\\hline
CMGP on outcome 1 &     5.20 ±0.25 &   45.71 ±2.80 &    15.27 ±2.77 &     2.48 ±0.11  &    14.74 ±0.74 &    96.80 ±0.94 &    8.03 ±6.20 &    84.02 ±6.21 &                - &                - &                - \\
CMGP on outcome 2 &     5.23 ±0.19 &   13.28 ±5.54 &    48.92 ±5.85 &                - &                - &                - &   38.02 ±8.75 &    51.93 ±9.10 &     4.02 ±0.23  &    22.98 ±1.40 &    97.75 ±0.48 \\
RF on outcome 1 &     4.90 ±0.27 &   \textbf{48.56 ±2.57} &    \textbf{12.85 ±2.10} &     2.51 ±0.07  &    14.49 ±0.68 &    96.67 ±0.92 &   10.54 ±5.06 &    80.38 ±5.96 &                - &                - &                - \\
RF on outcome 2 &     5.20 ±0.21 &   12.32 ±4.82 &    49.89 ±4.96 &                - &                - &                - &   \textbf{41.08 ±7.71} &    \textbf{50.11 ±7.44} &     4.30 ±0.20 &    24.79 ±1.45 &    97.45 ±0.88 \\
\bottomrule
\end{tabular}

\setlength{\tabcolsep}{1pt}
\begin{tabular}{p{2cm}|p{1cm}p{1cm}p{1cm}|p{1cm}p{1cm}p{1cm}p{1cm}|p{1cm}p{1cm}p{1cm}p{1cm}|p{1cm}}
\toprule
\multicolumn{12}{l|}{Jointly Partitioned (MOP-JCI)} \\\hline
&  & & & \multicolumn{4}{c|}{Outcome 1} & \multicolumn{4}{c|}{Outcome 2} & \\\hline
&     Num groups & Split Acc &          Split Err &       $V_{across}$ &        $V_{within}$ &            PEHE &               CI Width &       $V_{across}$ &        $V_{within}$ &            PEHE &               CI Width &     Cov (joint) \\ \hline
CMGP (SCR) &    5.20 ±0.21 &       100\% &        53\% &   46.44 ±2.26 &    14.91 ±2.33 &     2.42 ±0.08 &    13.69 ±0.43 &   59.23 ±5.95 &    33.33 ±5.19 &     4.19 ±0.30 &    23.94 ±1.58 &    97.17 ±0.63 \\
RF (SCR) &    5.17 ±0.22 &       100\% &        70\% &   47.68 ±1.80 &    12.96 ±1.16 &     2.45 ±0.07 &    13.51 ±0.69 &   53.57 ±4.57 &    38.03 ±4.27 &     4.42 ±0.18 &    25.88 ±1.27 &    97.13 ±0.66 \\
QRF (SCQR) &    4.90 ±0.23 &       100\% &        20\% &   \textbf{48.88 ±1.13} &    \textbf{10.92 ±0.64} &     2.62 ±0.09 &    15.54 ±1.03 &   \textbf{62.58 ±3.86} &    \textbf{26.43 ±2.86} &     4.40 ±0.16 &    28.51 ±1.56 &    97.13 ±0.98 \\
\bottomrule
\end{tabular}
\caption{Results from heteroscedastic synthetic data. We take the mean and standard deviation of each metric across 30 runs. Num groups is the number of subgroups generated. Best performance for $V_{across}$ and $V_{within}$ in each column are in bold.}
\label{tab:synth_hetsked}
\end{table*}

\newpage
\onecolumn
\twocolumn

\section{Subgroup Characteristics}\label{app:subgroup_chars}
In this section, the characteristics of the subgroups formed by each partioning method on each dataset are shown. For the synthetic datasets, we show the statistics of the age and time variable since those covariates determine the outcome distribution. For the semi-synthetic dataset (IHDP), we show the statistics of five of the continuous covariates in the dataset the birthweight (bw), neonatal health (nnhealth), weeks born preterm (preterm), age of mother (momage), and birth head size (bhead).  

\subsubsection{Synthetic data}

Table \ref{tab:uncorr_subgroup_separate} shows an example of the subgroup characteristics for the separate partitioning methods. Table \ref{tab:uncorr_subgroup} shows an example of the subgroup characteristics for the joint partitioning methods. Note the uncorrelated outcomes dataset was used in the paper as the synthetic dataset.

\begin{table*}[htbp!]
\vspace{-3.5em}
\scriptsize
\centering
\setlength{\tabcolsep}{1.5pt}

\begin{tabular}{ll}
\begin{tabular}{l|c|c|c|c|c|c|c}
\toprule

& \multicolumn{7}{c}{SCR CMGP separate on outcome 1}\\
Subgroup  &  count &  ALT mean &  ALT std &  time mean &  time std &  tau 1 mean &  tau 1 std \\
\midrule
1.0    &   49.0 &     16.03 &     5.51 &      13.66 &      0.83 &       -1.25 &       0.24 \\
2.0    &   27.0 &     15.18 &     4.90 &      11.45 &      0.36 &       -2.56 &       0.50 \\
3.0    &   29.0 &     15.31 &     5.70 &      10.16 &      0.47 &       -5.60 &       1.63 \\
4.0    &   28.0 &     15.25 &     5.40 &       8.33 &      0.43 &      -13.10 &       1.79 \\
5.0    &   67.0 &     16.34 &     5.42 &       5.71 &      0.95 &      -18.92 &       0.93 \\
\bottomrule

\end{tabular}
&\\
\begin{tabular}{l|c|c|c|c|c|c|c}
\toprule
& \multicolumn{7}{c}{SCR CMGP separate on outcome 2}\\
Subgroup  &  count &  ALT mean &  ALT std &  time mean &  time std &  tau 1 mean &  tau 1 std \\
\midrule
1.0    &   60.0 &     21.99 &     2.48 &       9.41 &      3.28 &       20.50 &       2.63 \\
2.0    &   31.0 &     17.72 &     0.52 &      10.03 &      3.44 &       13.60 &       1.66 \\
3.0    &   38.0 &     15.09 &     0.88 &       9.13 &      3.14 &        4.39 &       2.58 \\
4.0    &   34.0 &     10.01 &     3.14 &       8.84 &      2.92 &        0.16 &       0.22 \\
5.0    &   37.0 &     10.22 &     2.94 &       9.89 &      3.33 &        0.21 &       0.25 \\
\bottomrule
\end{tabular}
\end{tabular}

\begin{tabular}{ll}
\begin{tabular}{l|c|c|c|c|c|c|c}
\toprule
& \multicolumn{7}{c}{SCR RF separate on outcome 1}\\
Subgroup  &  count &  ALT mean &  ALT std &  time mean &  time std &  tau 1 mean &  tau 1 std \\
\midrule
1.0    &   46.0 &     16.26 &     6.00 &      12.95 &      1.19 &       -1.61 &       0.61 \\
2.0    &   31.0 &     16.76 &     6.40 &      12.65 &      1.19 &       -1.85 &       0.74 \\
3.0    &   23.0 &     16.26 &     4.98 &      10.00 &      0.44 &       -6.24 &       1.62 \\
4.0    &   45.0 &     15.33 &     5.20 &       7.85 &      0.66 &      -15.05 &       2.37 \\
5.0    &   25.0 &     15.87 &     5.63 &       5.64 &      0.87 &      -19.10 &       0.63 \\
6.0    &   30.0 &     14.77 &     5.82 &       5.09 &      0.84 &      -19.45 &       0.49 \\
\bottomrule
\end{tabular}
&\\
\begin{tabular}{l|c|c|c|c|c|c|c}
\toprule
& \multicolumn{7}{c}{SCR RF separate on outcome 2}\\
Subgroup  &  count &  ALT mean &  ALT std &  time mean &  time std &  tau 1 mean &  tau 1 std \\
\midrule
1.0    &   46.0 &     23.55 &     2.97 &       9.72 &      3.31 &       21.93 &       2.95 \\
2.0    &   27.0 &     17.65 &     1.26 &       8.89 &      2.67 &       12.72 &       3.83 \\
3.0    &   29.0 &     17.65 &     1.12 &       9.12 &      3.42 &       12.82 &       3.51 \\
4.0    &   72.0 &     11.38 &     3.08 &      10.79 &      2.59 &        0.93 &       1.35 \\
5.0    &   26.0 &     10.78 &     3.16 &       5.25 &      0.88 &        0.62 &       0.98 \\
\bottomrule
\end{tabular}
\end{tabular}
\caption{Subgroup characteristics from separate partitioning on synthetic data (Baseline R2P).}
\label{tab:uncorr_subgroup_separate}
\end{table*}

\begin{table*}[htbp!]
\vspace{-2em}
\scriptsize
\centering
\setlength{\tabcolsep}{1.5pt}

\begin{tabular}{l|c|c|c|c|c|c|c|c|c}
\toprule
& \multicolumn{7}{c}{SCR CMGP joint}\\
Subgroup &  count &  ALT mean &  ALT std &  time mean &  time std &  tau 0 mean &  tau 0std &  tau 1 mean &  tau 1 std \\
\midrule
0 &   60.0 &     19.77 &     3.18 &      12.08 &      1.61 &       -2.98 &       2.26 &       16.62 &       5.60 \\
1 &   47.0 &     19.94 &     2.92 &       6.89 &      1.67 &      -16.41 &       3.56 &       17.32 &       4.97 \\
2 &   54.0 &     11.45 &     3.10 &      12.19 &      1.74 &       -3.15 &       2.71 &        0.88 &       1.30 \\
3 &   39.0 &     11.71 &     2.70 &       6.26 &      1.37 &      -18.10 &       2.17 &        0.92 &       1.35 \\
\bottomrule
\end{tabular}

\begin{tabular}{l|c|c|c|c|c|c|c|c|c}
\toprule
& \multicolumn{7}{c}{SCR RF joint}\\
Subgroup &  count &  ALT mean &  ALT std &  time mean &  time std &  tau 0 mean &  tau 0std &  tau 1 mean &  tau 1 std \\

\midrule
0 &   72.0 &     20.10 &     3.20 &      11.94 &      2.03 &       -4.42 &       4.25 &       17.83 &       4.50 \\
1 &   28.0 &     19.77 &     2.69 &       6.31 &      1.11 &      -18.67 &       1.27 &       17.51 &       4.13 \\
2 &   32.0 &     11.93 &     2.60 &      13.61 &      1.04 &       -1.48 &       0.49 &        1.68 &       2.71 \\
3 &   27.0 &     10.74 &     4.33 &      10.22 &      0.75 &       -7.12 &       2.85 &        1.63 &       2.46 \\
4 &   41.0 &     11.65 &     3.45 &       6.49 &      1.49 &      -17.93 &       2.37 &        2.10 &       3.02 \\
\bottomrule

\end{tabular}
\begin{tabular}{l|c|c|c|c|c|c|c|c|c}
\toprule
& \multicolumn{7}{c}{SCQR RF joint}\\
Subgroup &  count &  ALT mean &  ALT std &  time mean &  time std &  tau 0 mean &  tau 0std &  tau 1 mean &  tau 1 std \\
\midrule
0 &   39.0 &     21.19 &     3.46 &      13.19 &      0.91 &       -1.47 &       0.45 &       18.58 &       4.72 \\
1 &   17.0 &     20.09 &     3.50 &      10.05 &      0.75 &       -6.44 &       2.60 &       16.92 &       4.86 \\
2 &   45.0 &     20.78 &     2.59 &       6.44 &      1.45 &      -17.50 &       2.70 &       18.40 &       3.69 \\
3 &   51.0 &     11.83 &     3.40 &      12.36 &      1.72 &       -2.79 &       2.56 &        1.74 &       2.76 \\
4 &   48.0 &     11.65 &     4.45 &       6.81 &      1.63 &      -16.50 &       3.40 &        1.85 &       2.52 \\
\bottomrule

\end{tabular}
\caption{Subgroup characteristics from joint partitioning on synthetic data (MOP-JCI).}
\label{tab:uncorr_subgroup}
\vspace{-2.5em}
\end{table*}

\subsubsection{Correlated Covariates}

Table \ref{tab:corr_cov_subgroup_separate} shows the subgroup characteristics for separately partitioned, and Table \ref{tab:corr_cov_subgroup} shows the subgroup characteristics when jointly partitioned.

\begin{table*}[htbp!]
\vspace{-2.5em}
\scriptsize
\centering
\setlength{\tabcolsep}{1.5pt}
\begin{tabular}{l|l}

\begin{tabular}{l|c|c|c|c|c|c|c}
\toprule
& \multicolumn{7}{c}{SCR CMGP separate on outcome 1}\\
Subgroup  &  count &  ALT mean &  ALT std &  time mean &  time std &  tau 1 mean &  tau 1 std \\

\midrule
1.0    &   33.0 &     19.17 &     1.47 &      12.48 &      0.84 &       -1.79 &       0.49 \\
2.0    &   30.0 &     17.45 &     1.74 &      10.80 &      0.32 &       -3.88 &       0.77 \\
3.0    &   42.0 &     17.12 &     1.79 &       9.65 &      0.43 &       -7.86 &       1.82 \\
4.0    &   46.0 &     15.64 &     1.45 &       8.29 &      0.43 &      -13.74 &       1.74 \\
5.0    &   30.0 &     13.19 &     2.19 &       5.45 &      1.72 &      -18.79 &       1.08 \\
6.0    &   19.0 &     13.55 &     1.70 &       6.00 &      1.41 &      -18.53 &       1.04 \\
\bottomrule

\end{tabular}
&\\
\begin{tabular}{l|c|c|c|c|c|c|c}
\toprule

& \multicolumn{7}{c}{SCR CMGP separate on outcome 2}\\
Subgroup  &  count &  ALT mean &  ALT std &  time mean &  time std &  tau 1 mean &  tau 1 std \\
\midrule
1.0    &   41.0 &     19.69 &     0.83 &      11.69 &      1.31 &       17.98 &       1.14 \\
2.0    &   49.0 &     17.62 &     0.50 &      10.14 &      1.35 &       13.38 &       1.64 \\
3.0    &   40.0 &     14.47 &     1.76 &       7.88 &      2.18 &        3.62 &       2.76 \\
4.0    &   27.0 &     14.57 &     1.60 &       7.64 &      1.67 &        3.76 &       2.96 \\
5.0    &   43.0 &     14.05 &     1.90 &       7.03 &      2.23 &        3.12 &       2.98 \\
\bottomrule
\end{tabular}
\end{tabular}

\begin{tabular}{ll}
\begin{tabular}{l|c|c|c|c|c|c|c}
\toprule
& \multicolumn{7}{c}{SCR RF separate on outcome 1}\\
Subgroup  &  count &  ALT mean &  ALT std &  time mean &  time std &  tau 1 mean &  tau 1 std \\
\midrule
1.0    &   26.0 &     19.96 &     1.10 &      12.18 &      1.25 &       -2.13 &       1.04 \\
2.0    &   25.0 &     16.91 &     1.00 &      11.24 &      0.92 &       -3.04 &       1.28 \\
3.0    &   46.0 &     16.38 &     1.67 &       9.52 &      0.33 &       -7.75 &       1.45 \\
4.0    &   40.0 &     15.01 &     1.56 &       8.17 &      0.44 &      -13.91 &       1.88 \\
5.0    &   33.0 &     14.35 &     1.57 &       6.88 &      0.38 &      -17.84 &       0.76 \\
6.0    &   30.0 &     12.79 &     1.67 &       5.17 &      1.15 &      -19.47 &       0.34 \\
\bottomrule
\end{tabular}
&\\
\begin{tabular}{l|c|c|c|c|c|c|c}
\toprule
& \multicolumn{7}{c}{SCR RF separate on outcome 2}\\
Subgroup  &  count &  ALT mean &  ALT std &  time mean &  time std &  tau 1 mean &  tau 1 std \\
\midrule
1.0    &   32.0 &     19.90 &     1.04 &      11.72 &      1.50 &       18.62 &       1.39 \\
2.0    &   47.0 &     17.27 &     0.49 &       9.83 &      1.57 &       13.02 &       1.63 \\
3.0    &   57.0 &     15.38 &     0.49 &       8.53 &      1.50 &        5.54 &       1.86 \\
4.0    &   64.0 &     12.94 &     1.02 &       6.59 &      1.63 &        0.83 &       0.60 \\
\bottomrule
\end{tabular}
\end{tabular}
\caption{Subgroup characteristics from separate partitioning on correlated covariate synthetic data (Baseline R2P).}
\label{tab:corr_cov_subgroup_separate}
\vspace{-2.5em}
\end{table*}

\begin{table*}[htbp!]
\scriptsize
\centering
\setlength{\tabcolsep}{1.5pt}
\begin{tabular}{l|c|c|c|c|c|c|c|c|c}
\toprule
& \multicolumn{7}{c}{SCR CMGP joint}\\
Subgroup &  count &  ALT mean &  ALT std &  time mean &  time std &  tau 0 mean &  tau 0std &  tau 1 mean &  tau 1 std \\

\midrule
0 &   32.0 &     19.58 &     1.13 &      11.87 &      1.42 &       -2.77 &       1.95 &       18.22 &       1.50 \\
1 &   54.0 &     17.42 &     0.47 &      10.44 &      1.25 &       -5.49 &       3.23 &       13.38 &       1.67 \\
2 &   24.0 &     15.41 &     0.82 &       9.82 &      0.97 &       -7.18 &       3.22 &        5.54 &       2.50 \\
3 &   42.0 &     16.07 &     0.86 &       7.56 &      0.90 &      -15.42 &       2.29 &        7.87 &       3.57 \\
4 &   48.0 &     13.09 &     1.18 &       5.85 &      1.74 &      -17.99 &       2.21 &        0.83 &       0.66 \\
\bottomrule
\end{tabular}

\begin{tabular}{l|c|c|c|c|c|c|c|c|c}
\toprule
& \multicolumn{7}{c}{SCR RF joint}\\
Subgroup &  count &  ALT mean &  ALT std &  time mean &  time std &  tau 0 mean &  tau 0std &  tau 1 mean &  tau 1 std \\

\midrule
0 &   33.0 &     20.31 &     1.26 &      12.36 &      1.52 &       -2.35 &       1.46 &       18.44 &       1.42 \\
1 &   51.0 &     17.53 &     0.54 &      10.34 &      1.72 &       -6.65 &       4.33 &       13.05 &       1.65 \\
2 &   29.0 &     15.37 &     0.89 &      10.06 &      0.72 &       -6.38 &       2.16 &        5.59 &       2.42 \\
3 &   21.0 &     15.02 &     0.96 &       8.50 &      0.33 &      -12.65 &       1.41 &        4.60 &       2.69 \\
4 &   66.0 &     13.24 &     1.89 &       6.27 &      1.45 &      -17.92 &       1.61 &        2.12 &       2.55 \\
\bottomrule
\end{tabular}

\begin{tabular}{l|c|c|c|c|c|c|c|c|c}
\toprule
& \multicolumn{7}{c}{SCQR RF joint}\\
Subgroup &  count &  ALT mean &  ALT std &  time mean &  time std &  tau 0 mean &  tau 0std &  tau 1 mean &  tau 1 std \\

\midrule
0 &   36.0 &     19.18 &     1.50 &      11.74 &      1.84 &       -3.32 &       2.39 &       17.37 &       2.28 \\
1 &   58.0 &     15.80 &     0.95 &      10.15 &      0.85 &       -5.94 &       2.57 &        7.44 &       3.59 \\
2 &   54.0 &     15.99 &     1.00 &       7.82 &      0.81 &      -14.94 &       2.28 &        7.94 &       3.90 \\
3 &   52.0 &     12.80 &     1.07 &       6.05 &      1.51 &      -18.10 &       2.03 &        0.76 &       0.67 \\
\bottomrule

\bottomrule
\end{tabular}
\caption{Subgroup characteristics from joint partitioning on correlated covariate synthetic data (MOP-JCI).}
\label{tab:corr_cov_subgroup}
\vspace{-2.5em}
\end{table*}

\subsubsection{Heteroscedasticity}

Table \ref{tab:hetsked_subgroup_separate} shows an example of the subgroup characteristics for the separate partitioning methods, and Table \ref{tab:hetsked_subgroup} shows an example of the subgroup characteristics for the joint partitioning methods.

\begin{table*}[t!]
\vspace{-3em}
\scriptsize
\centering
\setlength{\tabcolsep}{1.5pt}

\begin{tabular}{l|c|c|c|c|c|c|c}
\toprule
& \multicolumn{7}{c}{SCR CMGP separate on outcome 1}\\
Subgroup  &  count &  ALT mean &  ALT std &  time mean &  time std &  tau 1 mean &  tau 1 std \\
\midrule
1.0    &   37.0 &     18.04 &     5.31 &      11.51 &      1.95 &       -5.06 &       4.76 \\
2.0    &   23.0 &     16.46 &     4.42 &      11.59 &      2.41 &       -5.74 &       5.92 \\
3.0    &   20.0 &     16.71 &     4.41 &      11.57 &      2.19 &       -4.21 &       5.41 \\
4.0    &   55.0 &     16.30 &     4.69 &      11.39 &      2.29 &       -5.98 &       5.68 \\
5.0    &   37.0 &     14.57 &     4.81 &       5.98 &      1.17 &      -18.57 &       1.53 \\
6.0    &   28.0 &     17.25 &     6.45 &       5.85 &      1.05 &      -18.91 &       1.46 \\
\bottomrule
\end{tabular}

\begin{tabular}{l|c|c|c|c|c|c|c}
\toprule
& \multicolumn{7}{c}{SCR CMGP separate on outcome 2}\\
Subgroup  &  count &  ALT mean &  ALT std &  time mean &  time std &  tau 1 mean &  tau 1 std \\
\midrule
1.0    &   64.0 &     19.36 &     3.13 &       9.88 &      2.88 &       15.76 &       6.74 \\
2.0    &   48.0 &     20.85 &     3.71 &       9.71 &      3.26 &       18.16 &       7.75 \\
3.0    &   31.0 &     12.58 &     2.79 &      10.05 &      3.59 &        1.20 &       3.10 \\
4.0    &   16.0 &     11.74 &     3.32 &       8.84 &      3.35 &        0.73 &       2.08 \\
5.0    &   41.0 &     11.74 &     2.90 &       9.37 &      3.44 &        1.55 &       3.20 \\
\bottomrule
\end{tabular}

\begin{tabular}{l|c|c|c|c|c|c|c}
\toprule
& \multicolumn{7}{c}{SCR RF separate on outcome 1}\\
Subgroup  &  count &  ALT mean &  ALT std &  time mean &  time std &  tau 1 mean &  tau 1 std \\
\midrule
1.0    &   50.0 &     16.65 &     4.94 &      11.85 &      1.52 &       -2.94 &       3.50 \\
2.0    &   60.0 &     17.53 &     4.66 &      12.45 &      1.72 &       -2.52 &       3.52 \\
3.0    &   43.0 &     16.43 &     5.26 &       6.77 &      1.54 &      -17.09 &       3.60 \\
4.0    &   47.0 &     16.15 &     5.04 &       6.97 &      1.66 &      -16.43 &       3.84 \\
\bottomrule
\end{tabular}

\begin{tabular}{l|c|c|c|c|c|c|c}
\toprule
& \multicolumn{7}{c}{SCR RF separate on outcome 2}\\
Subgroup  &  count &  ALT mean &  ALT std &  time mean &  time std &  tau 1 mean &  tau 1 std \\
\midrule
1.0    &   42.0 &     22.54 &     2.28 &      11.63 &      1.96 &       22.12 &       4.94 \\
2.0    &   14.0 &     23.64 &     2.35 &       5.83 &      1.20 &       23.67 &       5.81 \\
3.0    &   25.0 &     16.57 &     1.94 &       9.46 &      2.69 &        9.77 &       7.00 \\
4.0    &   65.0 &     16.73 &     1.59 &       9.90 &      3.18 &        9.81 &       6.67 \\
5.0    &   30.0 &     10.55 &     2.12 &       9.63 &      3.58 &        0.10 &       1.57 \\
6.0    &   24.0 &     10.58 &     1.56 &       9.14 &      2.74 &       -0.26 &       1.26 \\
\bottomrule
\end{tabular}
\caption{Subgroup characteristics from separate partitioning on heteroscedastic synthetic data (Baseline R2P).}
\label{tab:hetsked_subgroup_separate}
\vspace{-2em}
\end{table*}

\begin{table*}[htbp!]
\scriptsize
\centering
\setlength{\tabcolsep}{1.5pt}
\begin{tabular}{l|c|c|c|c|c|c|c|c|c}
\toprule
& \multicolumn{7}{c}{SCR CMGP joint}\\
Subgroup &  count &  ALT mean &  ALT std &  time mean &  time std &  tau0 mean &  tau 0 std &  tau 1 mean &  tau1 std \\

\midrule
0 &   30.0 &     18.66 &     4.61 &      12.96 &      1.37 &       -2.00 &       4.24 &       11.90 &      11.47 \\
1 &   25.0 &     19.01 &     4.08 &      12.66 &      1.35 &       -3.05 &       3.91 &       13.50 &       9.59 \\
2 &   19.0 &     17.85 &     3.44 &       9.28 &      0.65 &       -8.75 &       2.84 &       11.75 &       9.59 \\
3 &   36.0 &     10.12 &     2.07 &      10.94 &      1.86 &       -5.45 &       3.82 &        0.33 &       2.75 \\
4 &   43.0 &     16.81 &     4.20 &       5.86 &      1.31 &      -18.31 &       1.81 &       11.18 &       8.84 \\
5 &   47.0 &     15.75 &     5.72 &       5.81 &      1.33 &      -18.22 &       2.04 &        9.69 &       8.95 \\
\bottomrule
\end{tabular}

\begin{tabular}{l|c|c|c|c|c|c|c|c|c}
\toprule
& \multicolumn{7}{c}{SCR RF joint}\\
Subgroup &  count &  ALT mean &  ALT std &  time mean &  time std &  tau0 mean &  tau 0 std &  tau 1 mean &  tau1 std \\

\midrule
0 &   26.0 &     19.89 &     2.88 &      10.89 &      1.69 &       -5.19 &       4.77 &       17.40 &       5.58 \\
1 &   26.0 &     21.46 &     3.74 &      10.35 &      1.61 &       -6.10 &       4.20 &       20.07 &       6.64 \\
2 &   70.0 &     12.11 &     3.08 &      11.21 &      1.73 &       -4.67 &       4.67 &        2.13 &       3.41 \\
3 &   37.0 &     20.46 &     2.76 &       6.03 &      1.21 &      -18.48 &       1.79 &       18.23 &       6.02 \\
4 &   41.0 &     11.28 &     3.29 &       6.01 &      1.10 &      -18.32 &       2.08 &        1.25 &       3.57 \\
\bottomrule
\end{tabular}

\begin{tabular}{l|c|c|c|c|c|c|c|c|c}
\toprule
& \multicolumn{7}{c}{SCQR RF joint}\\
Subgroup &  count &  ALT mean &  ALT std &  time mean &  time std &  tau0 mean &  tau 0 std &  tau 1 mean &  tau1 std \\
\midrule
0 &   51.0 &     20.62 &     3.11 &      11.66 &      1.91 &       -4.22 &       4.50 &       17.45 &       5.71 \\
1 &   40.0 &     13.21 &     2.25 &      13.01 &      1.32 &       -2.45 &       3.33 &        1.54 &       3.66 \\
2 &   22.0 &     12.16 &     2.93 &       9.79 &      0.64 &       -6.40 &       3.12 &        1.70 &       3.14 \\
3 &   41.0 &     20.44 &     2.93 &       6.37 &      1.43 &      -17.40 &       2.59 &       18.06 &       6.02 \\
4 &   46.0 &     13.35 &     2.69 &       5.91 &      1.32 &      -18.08 &       2.63 &        2.97 &       4.09 \\
\bottomrule
\end{tabular}
\caption{Subgroup characteristics from joint partitioning on heteroscedastic synthetic data (MOP-JCI).}
\label{tab:hetsked_subgroup}
\vspace{-5em}
\end{table*}

\subsubsection{Semi-synthetic data}

Table \ref{tab:IHDP_subgroup_separate} shows an example of the subgroup characteristics for the separate partitioning methods. Table \ref{tab:IHDP_subgroup} shows an example of the subgroup characteristics for the joint partitioning methods. 

\begin{table*}[htbp!]
\scriptsize
\centering
\setlength{\tabcolsep}{1.5pt}
\begin{tabular}{l|c|c|c|c|c|c|c}
\toprule
& \multicolumn{7}{c}{SCR CMGP separate on outcome 1}\\
Subgroup &  count  &  nnhealth mean &  nnhealth std &  momage mean &  momage std & tau 1 mean &  tau 1 std \\

\midrule
1.0    &   18.0 &           0.65 &          0.26 &         1.25 &        0.65 &       16.92 &       6.15 \\
2.0    &   61.0 &           0.81 &          0.50 &        -0.71 &        0.56 &       17.17 &       6.61 \\
3.0    &   35.0 &          -0.06 &          0.17 &         0.05 &        0.87 &       12.43 &       4.05 \\
4.0    &   11.0 &          -1.05 &          0.56 &         0.63 &        1.11 &        9.05 &       3.34 \\
5.0    &   25.0 &          -1.20 &          0.70 &        -0.04 &        0.79 &        8.48 &       3.56 \\
\bottomrule
\end{tabular}

\begin{tabular}{l|c|c|c|c|c|c|c}
\toprule
& \multicolumn{7}{c}{SCR CMGP separate on outcome 2}\\
Subgroup &  count & nnhealth mean &  nnhealth std &   momage mean &  momage std &  tau 2 mean &  tau 2 std \\
\midrule
1.0    &   27.0 &          -0.21 &          1.03 &         1.36 &        0.52 &        2.80 &       0.67 \\
2.0    &   32.0 &          -0.06 &          0.88 &         0.19 &        0.27 &        2.14 &       0.43 \\
3.0    &   18.0 &          -0.02 &          0.90 &        -0.89 &        0.40 &        1.44 &       0.57 \\
4.0    &   73.0 &           0.35 &          0.85 &        -0.55 &        0.77 &        0.74 &       0.30 \\
\bottomrule
\end{tabular}

\begin{tabular}{l|c|c|c|c|c|c|c}
\toprule
& \multicolumn{7}{c}{SCR RF separate on outcome 1}\\
Subgroup &  count &  nnhealth mean &  nnhealth std &  momage mean &  momage std & tau 1 mean &  tau 1 std \\
\midrule
1.0    &   40.0 &           1.12 &          0.42 &         0.20 &        0.95 &       19.19 &       7.18 \\
2.0    &   87.0 &          -0.35 &          0.75 &        -0.12 &        0.91 &       11.83 &       4.40 \\
3.0    &   23.0 &          -0.47 &          0.70 &        -0.10 &        0.82 &        8.56 &       2.56 \\
\bottomrule
\end{tabular}

\begin{tabular}{l|c|c|c|c|c|c|c}
\toprule
& \multicolumn{7}{c}{SCR RF separate on outcome 2}\\
Subgroup &  count &  nnhealth mean &  nnhealth std &   momage mean &  momage std &   tau 2 mean &  tau 2 std \\
\midrule
1.0    &  150.0 &           0.02 &          0.94 &        -0.03 &        0.91 &         1.5 &       0.88 \\
\bottomrule
\end{tabular}
\caption{Subgroup characteristics from separate partitioning on IHDP data (Baseline R2P).}
\label{tab:IHDP_subgroup_separate}
\vspace{-1em}
\end{table*}

\begin{table*}[ht!]
\tiny
\centering
\setlength{\tabcolsep}{1.5pt}
\begin{tabular}{l|c|c|c|c|c|c|c|c|c}
\toprule
& \multicolumn{8}{c}{SCR CMGP joint}\\
Subgroup &  count   &  nnhealth mean &  nnhealth std &  momage mean &  momage std   &  tau 1 mean &  tau 1 std &  tau 2 mean &  tau 2 std \\
\midrule
0 &   33.0 &           0.77 &          0.37 &         0.24 &        0.91 &       19.47 &       6.00 &        2.12 &       0.61 \\
1 &   19.0 &          -0.23 &          0.23 &         0.21 &        0.86 &       13.13 &       3.67 &        2.14 &       0.69 \\
2 &   24.0 &          -1.62 &          0.93 &         0.63 &        0.88 &        8.41 &       3.70 &        2.34 &       0.83 \\
3 &   74.0 &           0.02 &          1.08 &        -0.38 &        0.91 &       12.00 &       5.16 &        0.82 &       0.38 \\
\bottomrule
\end{tabular}

\begin{tabular}{l|c|c|c|c|c|c|c|c|c}
\toprule
& \multicolumn{8}{c}{SCR RF joint}\\
Subgroup &  count &  nnhealth mean &  nnhealth std &  momage mean &  momage std &  tau 1 mean &  tau 1 std &  tau 2 mean &  tau 2 std \\
\midrule
0 &   19.0 &           1.06 &          0.46 &        -0.10 &        0.80 &       19.49 &       7.74 &        1.25 &       0.75 \\
1 &   38.0 &           0.99 &          0.42 &         0.14 &        1.25 &       19.24 &       5.40 &        1.80 &       1.12 \\
2 &   39.0 &          -0.71 &          0.91 &         0.77 &        0.71 &       10.52 &       6.00 &        2.02 &       0.83 \\
3 &   54.0 &          -0.26 &          0.56 &        -0.84 &        0.41 &       10.74 &       4.26 &        0.92 &       0.55 \\
\bottomrule
\end{tabular}

\begin{tabular}{l|c|c|c|c|c|c|c|c|c}
\toprule
& \multicolumn{8}{c}{SCQR RF joint}\\
Subgroup &  count &   nnhealth mean &  nnhealth std &  momage mean &  momage std &  tau 1 mean &  tau 1 std &  tau 2 mean &  tau 2 std \\
\midrule
0 &   54.0 &           0.98 &          0.36 &        -0.16 &        0.95 &       18.20 &       6.14 &        1.49 &       0.94 \\
1 &   40.0 &          -0.27 &          0.38 &         0.40 &        1.04 &       13.25 &       4.43 &        2.25 &       0.89 \\
2 &   32.0 &          -0.07 &          0.29 &        -0.60 &        0.82 &       11.17 &       3.98 &        0.78 &       0.35 \\
3 &   24.0 &          -1.63 &          0.81 &         0.13 &        1.14 &        7.57 &       2.65 &        1.60 &       0.93 \\
\bottomrule
\end{tabular}
\caption{Subgroup characteristics from joint partitioning on IHDP data (MOP-JCI).}
\label{tab:IHDP_subgroup}
\end{table*}

\end{document}